%% file: main_neurips.tex
\documentclass{article}

\PassOptionsToPackage{numbers, compress}{natbib}
 \usepackage[preprint]{neurips_2026}

\usepackage[utf8]{inputenc} 
\usepackage[T1]{fontenc}    
\usepackage{hyperref}       
\usepackage{url}            
\usepackage{booktabs}       
\usepackage{amsfonts}       
\usepackage{nicefrac}       
\usepackage{microtype}      
\usepackage{xcolor}         
\usepackage{subcaption}
\newcommand{\idea}[1]{}
\usepackage{wrapfig}
\usepackage{amsmath}
\usepackage{amssymb}
\usepackage{mathtools}
\usepackage{amsthm}
\usepackage{mathbbol}
\usepackage{bbm}
\usepackage{bm}

\usepackage{tikz}
\usepackage{algorithm}
\usepackage{algorithmic}
\usepackage{xcolor}
\usepackage{multirow}
\usepackage{wrapfig}
\usepackage{bbm}
\usepackage{tabularx}

\newcommand{\todo}[1]{
  \textbf{\textcolor{red}{TODO: #1}}
}

\idea{Latent Functional Non-Negative Matrix Factorization for Point Process Data}

\idea{Non-Negative Matrix Factorization for Point Processes}
\idea{NMF for Event Data}
\idea{NMF for Point Processes}
\title{Non-Negative Matrix Factorization for Event Data}

\hypersetup{
  colorlinks=true,
  linkcolor=blue,
  citecolor=blue,
  urlcolor=blue
}

\usepackage[capitalize,noabbrev]{cleveref}

\theoremstyle{plain}
\newtheorem{theorem}{Theorem}[section]
\newtheorem{proposition}[theorem]{Proposition}

\theoremstyle{definition}

\theoremstyle{remark}
\newtheorem{remark}[theorem]{Remark}

\input{preamble.tex}

\author{
  Rapha\"el Romero \\
  Ghent University \\
  Ghent, Belgium \\
  \texttt{raphael.romero@ugent.be}
}

\begin{document}

\maketitle


\begin{abstract}
  Continuous-time event data, in which entities emit instantaneous events over time, arises naturally across many domains such as neuroscience, seismology, and social networks. Non-negative matrix factorization (NMF) is a natural tool to uncover interpretable structure in such data, but it has so far only been applied after binning or smoothing the entity-level counting measures. This preprocessing step comes with the risk of erasing entity-level heterogeneities and fine-grained temporal features. In this paper, we introduce EventNMF, a continuous-time non-negative factorization model that operates directly on event times: each entity's events are modeled as a Poisson process whose intensity factorizes through a non-negative B-spline basis, and a simple estimation procedure recovers interpretable temporal templates shared across entities. The resulting method is mathematically principled, easy to implement, and computationally efficient. We further show that standard binned-count approaches arise as the special case of degree-zero splines, explore bias-variance tradeoffs and compare against existing methods on a synthetic latent factor model, and demonstrate the effectiveness of EventNMF on several real-world applications. A Python implementation is available at \url{https://anonymous.4open.science/r/fdna_public-14BC/}.
\end{abstract}

\section{Introduction}

Many real-world datasets may be viewed as a collection of event times occurring in a given observation interval. Examples include timestamps of user actions on a website, communication events on a social network \cite{safdariCommunityDetectionAnomaly2024}, neuronal spike trains \cite{cunninghamDimensionalityReductionLargescale2014}, seismic events \cite{ogataStatisticalModelStandard1989}, and single cell biology \cite{carstensenMultivariateHawkesProcess2010, gustoFADOStatisticalMethod2005, picardPCAPointProcesses2024}. This leads to datasets where each data point can be viewed as a collection of discrete timestamps in a time interval, or more generally as a collection $t_{i,1}, \ldots, t_{i,n_i}$ of points in a continuous space $\Tcal$, typically a time interval $\Tcal = [0,1]$. We refer to such data as \emph{event data}, and to the individual data points as \emph{entities}. Point processes have been widely used to model such data, and in particular Poisson processes are a common choice for modeling the generative process underlying event data \cite{ogataStatisticalModelStandard1989,daleyIntroductionTheoryPoint2005,zhangPoissonIntensityEstimation}. Different from standard vector datasets, event data is inherently continuous in time, and each data point can be viewed without information loss as a counting measure associated with its event times, namely $\Ybb_i = \sum_{j=1}^{n_i} \delta_{t_{i,j}}$ where $\delta_x$ is the Dirac delta measure at $x$. Finding interpretable temporal patterns in such datasets is a significant challenge across many fields. For instance, in neuroscience, dimensionality reduction methods are commonly used to extract low-dimensional representations of neural activity that capture the collective dynamics of large populations of neurons \cite{cunninghamDimensionalityReductionLargescale2014,LOPESDOSSANTOS2013149,onkenUsingMatrixTensor2016}. In seismology, identifying temporal patterns in earthquake catalogs can provide insights into underlying geophysical processes \cite{ogataStatisticalModelStandard1989,sequeiraBlindIntensityEstimation1997}. In network analysis, low-rank models applied to temporal communication data can uncover latent community structure and help detect significant changes in interaction patterns or anomalies over time \cite{safdariCommunityDetectionAnomaly2024,romeroMultiresolutionAnalysisStatistical2025,modellIntensityProfileProjection2023}. 

Nonnegative Matrix Factorization (NMF), as introduced in the seminal work of Lee and Seung \cite{leeLearningPartsObjects1999b}, appears as a natural candidate to detect such patterns. However, extensions of NMF to event data typically proceed by discretizing the data or smoothing it over time to obtain temporal signals that are used as input to standard dimensionality reduction methods \cite{leeAlgorithmsNonnegativeMatrix,ramsayFunctionalDataAnalysis2005,muellerFunctionalDataAnalysis2014,hautecoeurNonnegativeMatrixFactorization2020}. Another notable approach, perhaps the closest to our proposed method, consists in considering each data point as an inhomogeneous time series of counts and attempting to model these counts continuously over time using a Poisson-adapted version of NMF \cite{backenrothNonnegativeDecompositionFunctional2020}. Yet, binning or smoothing event data prior to applying NMF may lead to two types of issues. On the one hand, binning is known to fundamentally alter the statistical properties of the input point process signals themselves \cite{cessacMathematicalConsequencesBinning2017}. On the other hand, it introduces a bias-variance tradeoff in the choice of bin width or kernel bandwidth that is difficult to control \cite{shimazakiMethodSelectingBin2007,shimazakiKernelBandwidthOptimization2010}, which can mean either erasing useful variability between entities or smoothing out fine-grained temporal features in the estimated latent factors. 

To avoid these pitfalls, in this work we propose a novel latent factor model for event data based on the Poisson process likelihood. Our model assumes that each entity's event intensity is a non-negative linear combination of a small number of latent temporal factors, which are themselves decomposed on a non-negative B-spline basis. We derive efficient multiplicative updates for parameter estimation and evaluate the method on synthetic and real-world datasets.

\idea{
\ptitle{Challenges/Limitations of existing approaches}
- Binning or smoothing the data prior to applying NMF may lead to loss of information and introduce artifacts, especially if the bin size or smoothing parameters are not chosen appropriately.
- Philosophically In the presence of discrete events in continuous time, the best strategy following the Maximum Entropy principle is to model the latent factors generating these events as smoothly varying functions rather than histogram. This makes functional bases such as splines particularly relevant.
}

\idea{example structure used in a previous paper: Contributions. In Section 3, we formulate change detection as a statistical signal processing problem, where the goal is to recover edge-level temporal signals from noisy dynamic network observations. In Section 4, we present a new statistical method for multi-resolution change detection, supported by theoretical guarantees. In Section 5, we evaluate our method on both synthetic and realworld datasets, demonstrating that ANIE outperforms fixed-resolution approaches by effectively capturing changes at multiple time scales in dynamic networks.}

\ptitle{Outline}
The paper is organized as follows. Section~\ref{sec:related_work} connects the proposed method to existing work. Section~\ref{sec:method} provides background on nonnegative matrix factorization (NMF) and Poisson processes, and formally introduces the EventNMF model. Section~\ref{sec:simulations} evaluates the proposed model on a synthetic scenario with known latent factors and loading matrix. Section~\ref{sec:applications} explores three applications of EventNMF: a single-trial analysis of multi-electrode neuronal spike train recordings, a dataset of earthquake events, and a dataset of face-to-face contacts in a primary school.

\section{Related Work\label{sec:related_work}}

\input{related_work.tex}

\clearpage
\section{Non-Negative Matrix Factorization for Event Data \label{sec:method}}

\input{method.tex}

\input{experiments_synthetic.tex}

\section{Applications\label{sec:applications}}
\input{applications.tex}

  \section{Conclusion}
We have presented a novel latent factor model for event data based on Poisson process likelihoods. Our approach extends non-negative matrix factorization to continuous-time event data, providing an interpretable framework for discovering temporal patterns in large-scale event datasets. Crucially, our method operates directly on raw event times without requiring any binning or smoothing, and we demonstrate its applicability across a variety of domains.

Several directions for future work remain. The scalability of the method could be further improved through random subsampling of event times, which would make it applicable to very large datasets consisting of millions of entities and billions of events. Additionally, while our current implementation relies on multiplicative updates, exploring alternative optimization algorithms and characterizing their respective benefits and limitations in the context of event data represents a promising research avenue. Finally, an important open question concerns self-exciting processes: since Poisson processes are not self-exciting by construction, understanding the behavior of EventNMF on data generated by coupled multivariate Hawkes processes both theoretically and empirically would be a natural and valuable extension.

\begin{ack}
The research leading to these results has received funding from the Special Research Fund (BOF) of Ghent University (BOF20/IBF/117), from the Flemish Government under the ``Onderzoeksprogramma Artificiële Intelligentie (AI) Vlaanderen'' programme, from the FWO (project no. G0F9816N, 3G042220, G073924N). Funded by the European Union (ERC, VIGILIA, 101142229). Views and opinions expressed are however those of the author(s) only and do not necessarily reflect those of the European Union or the European Research Council Executive Agency. 
\end{ack}

\bibliographystyle{plainnat}
\bibliography{../references}

\appendix
\input{appendix.tex}



\end{document}

%% file: preamble.tex
\usepackage{tcolorbox}

\renewenvironment{align}{
    \begin{equation}
        \begin{aligned}
}{
        \end{aligned}
    \end{equation}
}
\renewenvironment{align*}{
    \begin{equation*}
        \begin{aligned}
}{
        \end{aligned}
    \end{equation*}
}

\DeclareMathOperator*{\argmin}{arg\,min}

\renewcommand{\epsilon}{\varepsilon}

\newcommand{\triplenorm}[1]{{\left\vert\kern-0.25ex\left\vert\kern-0.25ex\left\vert #1 
    \right\vert\kern-0.25ex\right\vert\kern-0.25ex\right\vert}}

\newcommand{\note}[1]{}

\newcommand{\ptitle}{\textbf}

\newcommand{\ind}{\mathbbm{1}}

\newcommand{\Ecal}{\mathcal{E}}

\newcommand{\pparentheses}[1]{\bigl( #1 \bigr)}

\newcommand{\brackets}[1]{\left[ #1 \right]}

\newcommand{\deltaequal}{\overset{\Delta}{=}}

\def\eqref#1{equation~\ref{#1}}

\def\1{\bm{1}}

\def\hbar{{\bar{h}}}

\DeclareMathAlphabet{\mathsfit}{\encodingdefault}{\sfdefault}{m}{sl}
\SetMathAlphabet{\mathsfit}{bold}{\encodingdefault}{\sfdefault}{bx}{n}

\def\Ecal{{\mathcal{E}}}

\def\Ical{{\mathcal{I}}}

\def\Lcal{{\mathcal{L}}}

\def\Tcal{{\mathcal{T}}}

\def\Xcal{{\mathcal{X}}}

\def\Uhat{{\hat{U}}}

\def\Ebb{{\mathbb{E}}}

\def\Pbb{{\mathbb{P}}}

\def\Rbb{{\mathbb{R}}}

\def\Ybb{{\mathbb{Y}}}

\def\hbar{{\bar{h}}}

%% file: related_work.tex
The proposed EventNMFmethod draws on several lines of research, including nonnegative matrix factorization (NMF), functional data analysis (FDA), and point process modeling. We briefly review the most relevant works below.

Nonnegative matrix factorization (NMF) is a foundational dimensionality reduction technique \cite{leeAlgorithmsNonnegativeMatrix} with applications in signal processing \cite{smaragdisNonnegativeMatrixFactorization2003}, recommender systems \cite{korenMatrixFactorizationTechniques2009}, and neuroscience \cite{cunninghamDimensionalityReductionLargescale2014,LOPESDOSSANTOS2013149,mackeviciusUnsupervisedDiscoveryTemporal2019a}. Recent work has extended NMF to functional and longitudinal data by constraining latent factors to smooth B-spline bases. Such approaches have been applied to continuous signals sampled at irregular time points \cite{zdunekBSplineSmoothingFeature2014,hautecoeurNonnegativeMatrixFactorization2020}, temporal count data using Poisson likelihoods \cite{backenrothNonnegativeDecompositionFunctional2020}, and continuous tensor decompositions \cite{sortLatentFunctionalPARAFAC2024}. To our knowledge, however, NMF has not previously been studied in the context of point process data.

Functional data analysis studies collections of smooth signals observed over continuous domains, often on irregular grids \cite{ramsayFunctionalDataAnalysis2005}. A central tool in this setting is functional principal component analysis (FPCA), which represents data through low-dimensional smooth modes of variation. Extensions to multivariate functional data observed on heterogeneous domains have also been developed \cite{happMultivariateFunctionalPrincipal2018}. These ideas have recently been adapted to point process data. In particular, entity-specific intensities have been modeled as smooth random functions, with shared functional structure used to improve estimation in sparse-event settings \cite{muellerFunctionalDataAnalysis2014}. More recently, FPCA formulations for point processes based on random measure representations have been proposed \cite{picardPCAPointProcesses2024}.

Point processes provide a natural framework for modeling event data occurring in continuous time \cite{daleyIntroductionTheoryPoint2005,kingman1992poisson}. Among the most widely used models are inhomogeneous Poisson processes, where events occur independently according to a time-varying intensity function, and Hawkes processes, which allow past events to influence future occurrences \cite{ogataStatisticalModelStandard1989,carstensenMultivariateHawkesProcess2010,gustoFADOStatisticalMethod2005}. A central problem in this literature is nonparametric intensity estimation from observed event times. Existing approaches include penalized projection estimators \cite{reynaud-bouretAdaptiveEstimationIntensity2003}, reproducing kernel methods \cite{flaxmanPoissonIntensityEstimation2017}, and wavelet-based estimators \cite{kolaczykWaveletShrinkageEstimation1999,talebMultiresolutionAnalysisPoint2021}. These methods have been widely applied in seismology \cite{ogataStatisticalModelStandard1989}, genomics \cite{gustoFADOStatisticalMethod2005,carstensenMultivariateHawkesProcess2010}, and neuroscience \cite{cunninghamMethodsEstimatingNeural2009}.

Our approach sits at the intersection of these lines of work. Like functional NMF methods, we seek a low-dimensional nonnegative decomposition of temporal structure across entities. At the same time, our model directly estimates continuous-time intensity functions from point process observations using a Poisson process likelihood. This connects dimensionality reduction and point process intensity estimation within a unified framework for event data.
\idea{
  Point Processes are 
  - seminal books, types of point processes (book by daley and vere-jones,)
  types of point processes Hawkes(ogataStatisticalModelStandard1989, carstensenMultivariateHawkesProcess2010
gustoFADOStatisticalMethod2005) Poisson processes ()

-intensity estimation : methodologies (... ) applications (... cunninghamMethodsEstimatingNeural2009)

}

%% file: method.tex
In this section we start by briefly introducing the notations used throughout the paper, and then we present our proposed model for event data, which we call EventNMF. The EventNMF objective is derived from the Poisson process likelihood, and we show that it can be optimized using efficient multiplicative updates, akin to those used in standard NMF.

\ptitle{Notations.}
In this paper we consider datasets $\Xcal = \{\Ecal_i\}_{i=1}^N$, where each “data point” $\Ecal_i = \{t_{i,m}\}_{m=1}^{M_i}$ is a set of event times associated with entity $i$, observed over a continuous time interval $\Tcal = [0,T]$.
 We can equivalently represent $\Ecal_i$ as a \textit{counting measure} $\Ybb_i=\sum_{m=1}^{M_i} \delta_{t_{i,m}}$, where $\delta_t$ is the Dirac measure at time $t$ and $M_i$ is the number of events for entity $i$. 
In the remainder of the paper, we use the following notation for vector operations: for two $R$-dimensional vectors $u$ and $v$, we denote by $u \odot v = (u_i v_i)_{i=1}^R$ their element-wise (Hadamard) product, by $u \otimes v = (u_i v_j)_{i,j=1}^R$ their outer product, and by $u^{\top} v = \sum_{i=1}^R u_i v_i$ their dot product.

\subsection{A Low-Rank Poisson Intensity Model\label{sec:model}}
\input{figure1.tex}

In EventNMF, each entity's associated counting measure $\Ybb_i$ is modeled 
as arising from a Poisson process with intensity function
$\lambda_i : [0,T] \to \Rbb_+$. We assume the intensity functions admit a 
low-rank factorization
\[
\lambda_i(t) = \sum_{r=1}^R u_{ir}\, f_r(t),
\quad u_{ir} \geq 0,\quad f_r(t) \geq 0,
\]
where $u_{ir}$ are nonnegative \emph{loadings} and $f_r$ are nonnegative
\emph{latent factors} shared across entities. To make the model parametric,
we expand each latent factor in a fixed nonnegative basis
$\{\phi_b\}_{b=1}^B$ such as B-splines. Specifically, we write
\[
  f_r(t) = \sum_{b=1}^B \gamma_{rb}\,\phi_b(t), \quad \gamma_{rb} \geq 0.
\]
Denoting $\Phi(t) = (\phi_1(t), \ldots, \phi_B(t))^\top \in \Rbb^B$,
$u_i = (u_{i1}, \ldots, u_{iR})^\top \in \Rbb_+^R$, and
$\Gamma = (\gamma_{rb}) \in \Rbb_+^{R \times B}$, the intensity functions
take the compact form $\lambda_i(t) = u_i^\top \Gamma\, \Phi(t)$.
The model is thus parametrized by the loading matrix
$U = (u_{ir}) \in \Rbb_+^{N \times R}$ and the factor coefficient matrix
$\Gamma \in \Rbb_+^{R \times B}$.

\ptitle{B-spline functions}

In this paper we use B-splines as the basis functions
$\{\phi_b\}_{b=1}^B$, since they form a flexible family of nonnegative
functions with local support that can approximate a wide range of
intensity-function shapes. The B-spline basis is defined recursively
over the degree. Given a knot sequence
$k_1 \le k_2 \le \cdots \le k_{B+p+1}$ and a degree $p \ge 0$, the
$b$-th B-spline of degree $p$ is defined by the Cox--de Boor recursion
\begin{align*}
  B_{b,p}(t) =
    \frac{t - k_b}{k_{b+p} - k_b}\, B_{b,\,p-1}(t)
    \;+\;
    \frac{k_{b+p+1} - t}{k_{b+p+1} - k_{b+1}}\, B_{b+1,\,p-1}(t),
\end{align*}
with the base case $p=0$ given by indicator functions on adjacent intervals, namely $B_{b,0} \deltaequal \ind_{[k_b, k_{b+1})}$.
The basis functions used in EventNMF are
$\phi_b(t) \deltaequal B_{b,p}(t)$ for $b = 1, \ldots, B$.
We adopt the standard convention that any term with a zero denominator
is set to zero. Intuitively, B-splines of degree $p$ are piecewise polynomials of degree
$p$ that are nonnegative and have local support over $p+2$ adjacent
knots. The choice of degree $p$ and number of basis functions $B$
controls the model capacity: higher degree and more knots
yield more expressive intensity functions, while lower degree and fewer
knots yield smoother ones.
\subsection{Estimation Procedure}
The parameters of the EventNMF model are estimated by minimizing the negative log-likelihood of the observed data under the Poisson process model defined above. 

\begin{proposition}[Negative log-likelihood of EventNMF]
  Under the Poisson process model defined in Sec~\ref{sec:model}, with parametric intensity $\lambda_i(t) = u_i^\top \Gamma\, \Phi(t)$, the negative log-likelihood of the observed counting measures $\{\Ybb_i\}$  is given by
\begin{align*}
  \Lcal(\{\Ybb_i\}; U, \Gamma) = \sum_{i=1}^N l_i(U, \Gamma), \quad l_i(U, \Gamma) = u_i^\top \Gamma\, I(\Phi) - \sum_{\tau \in \Ecal_i} \log\!\left(u_i^\top \Gamma\, \Phi(\tau)\right),
\end{align*}
where $I(\Phi) = \int_{\Tcal} \Phi(t)\,dt \in \Rbb^B$ is the vector whose entries are the integrals of the basis functions. 
\end{proposition}
This follows from the form of the Poisson process likelihood~\cite{daleyIntroductionTheoryPoint2005} and the linearity of the integral. We next derive the gradients of $l_i$, which will be used to optimize the loss function defined above.
The proof is given in Appendix~\ref{appendix:proof_gradients}.

\begin{proposition}[Gradients of EventNMF, \label{prop:gradients}]
  The gradients of $l_i$ with respect to $u_i$ and $\Gamma$ are
  \begin{align*}
    \nabla_{u_i} \Lcal(U,\Gamma)
     & =\nabla_{u_i} l_i(U,\Gamma)
    = \Gamma\, I(\Phi)
    - \sum_{\tau \in \Ecal_i}
    \frac{\Gamma\,\Phi(\tau)}{u_i^\top \Gamma\,\Phi(\tau)}, \\
    \nabla_{\Gamma} \Lcal(U,\Gamma)
     & =\sum_{i=1}^N \nabla_{\Gamma} l_i(U,\Gamma),\quad
    \nabla_{\Gamma} l_i(U,\Gamma) = u_i \otimes I(\Phi)
    - \sum_{\tau \in \Ecal_i}
    \frac{u_i \otimes \Phi(\tau)}{u_i^\top \Gamma\,\Phi(\tau)}.
  \end{align*}
\end{proposition}

\newcommand{\gpU}{g^{(U, +)}}
\newcommand{\gmU}{g^{(U, -)}}
\newcommand{\gpG}{g^{(\Gamma, +)}}
\newcommand{\gmG}{g^{(\Gamma, -)}}

\ptitle{Optimization} In order to fit the EventNMF model, we need to optimize the loss function $\Lcal({\Ybb_i}; U, \Gamma)$ with respect to the non-negative parameters $U$ and $\Gamma$. To do so, we derive efficient multiplicative updates inspired by the classical NMF algorithm \cite{leeAlgorithmsNonnegativeMatrix}, which guarantee that the parameters remain non-negative throughout the optimization process. Alternative approaches include projected gradient descent, coordinate descent, ALS (alternating least squares), and IALS (inexact alternating least squares), where at each update step the negative coefficients of the unconstrained solution are set to zero \cite{yuNONNEGATIVEMATRIXFACTORIZATION2014}; we focus on multiplicative updates here due to their simplicity and efficiency.

\idea{
\subsection{Properties}
  The vector giving, for each entity, the expected number of events under the fitted model is
  \begin{align*}
    \pparentheses{
      \Ebb\left[\Ybb_i([0,T])\right]
    }_{i=1}^N
    =
    \pparentheses{
      u_i^\top \Gamma\, I(\Phi)
    }_{i=1}^N    = U \Gamma\, I(\Phi).
  \end{align*}
\end{proposition}

\begin{proof}
  This follows from the definition of the Poisson process and the form of the intensity function. The expected number of events for entity $i$ is given by the integral of the intensity function over the time interval $[0,T]$:
  \begin{align*}
    \Ebb\left[\Ybb_i([0,T])\right]
     & = \int_0^T \lambda_i(t)\,dt              \\
     & = \int_0^T u_i^\top \Gamma\, \Phi(t)\,dt \\
     & = u_i^\top \Gamma \int_0^T \Phi(t)\,dt   \\
     & = u_i^\top \Gamma\, I(\Phi).
  \end{align*}
  Stacking these row-wise over $i = 1, \dots, N$ yields $U \Gamma\, I(\Phi)$.
\end{proof}

}

\ptitle{Computational complexity.}
Let $E = \sum_{i=1}^N |\mathcal{E}_i|$ denote the total number of events.
Before the optimization loop, EventNMF precomputes three quantities:
(i)~the $B$-vector of basis integrals $I(\Phi)=\int_0^T \Phi(t)\,dt$;
(ii)~the $B \times E$ matrix $\Phi_{\mathcal{E}}$, whose $e$-th column is $\Phi(\tau_e)$;
and (iii)~the sparse $N \times E$ event-indicator matrix $M$, with $M_{ie}=1$ if event $e$ belongs to entity $i$.
All three are computed once at cost $\mathcal{O}(BE)$ and reused across all iterations.

Each iteration is dominated by the product $\Gamma \Phi_{\mathcal{E}} \in \mathbb{R}^{R \times E}$, computed once per step and used to assemble all gradient terms.
The gradient with respect to $U$ reduces to a sparse multiply by $M$, and the gradient with respect to $\Gamma$ to a single dense multiply.
The per-iteration cost is $\mathcal{O}(RBE)$, linear in the number of events. In practice, EventNMF fits a dataset with $N=1{,}000$ entities and $E \approx 27{,}000$ events ($B=40$, $R=3$) in under 1 second on a standard CPU, and scales to $N=10{,}000$ entities and $E \approx 10^6$ events in under 1 minute (200 iterations).
\idea{

  Justification:
  We denote $\gpu_{ir}$ and $\gmu_{ir}$ as the positive and negative parts of the gradient with respect to $u_{ir}$, and similarly $\gpG_{rb}$ and $\gmG_{rb}$ for the gradient with respect to $\gamma_{rb}$.

  For $U$, the gradient depends only on neuron $i$'s events:
  \begin{align*}
    \gpu_{ir} & =  \sum_{b=1}^B \gamma_{rb}I(\phi_b),                                                   \\
    \gmu_{ir} & = \sum_{\tau \in \Ecal_i} \frac{\sum_{b=1}^B \gamma_{rb}\phi_b(\tau)}{\lambda_i(\tau)},
  \end{align*}
  where $\lambda_i(\tau) = \sum_{s=1}^R U_{is} \sum_{c=1}^B \gamma_{sc}\psi_{c}(\tau)$ is the intensity at time $\tau$ for neuron $i$.

  For $\Gamma$, the gradient sums over all neurons:
  \begin{align*}
    \gpG_{rb} & = \sum_{i=1}^N u_{ir} I(\phi_b),                                                    \\
    \gmG_{rb} & = \sum_{i=1}^N u_{ir} \sum_{\tau \in \Ecal_i} \frac{\phi_b(\tau)}{\lambda_i(\tau)}.
  \end{align*}

  We note that all of those quantities are non-negative due to the non-negativity of $U$, $\Gamma$, and the basis functions $\phi_b(t)$. Similar to the classical NMF setting \cite{leeAlgorithmsNonnegativeMatrix}, we can derive multiplicative updates from these positive and negative gradient parts:

  \begin{align*}
    u_{ir}      & \leftarrow u_{ir} \frac{\gmu_{ir}}{\gpu_{ir}},      \\
    \gamma_{rb} & \leftarrow \gamma_{rb} \frac{\gmG_{rb}}{\gpG_{rb}}.
  \end{align*}

  It can be shown that these updates decrease the loss function at each iteration, similar to the classical NMF case \cite{leeAlgorithmsNonnegativeMatrix}.
}

\subsection{Remarks}




\ptitle{Connection to Poisson-NMF (degree-0 basis).}
When the B-spline degree is zero, the basis reduces to piecewise-constant indicators $\phi_b(t)=\mathbb{1}_{[k_b,k_{b+1})}(t)$, so that $\Phi(t)=e_b$ for $t\in[k_b,k_{b+1})$ where $e_b\in\Rbb^b$ is the $b$-th standard basis vector. Define the binned count matrix $X\in\mathbb{N}^{N\times B}$ with $X_{ib}=\Ybb_i([k_b,k_{b+1}))$ and expected counts $\hat{X}_{ib}=u_i^\top \Gamma e_b (k_{b+1}-k_b)$.

This result shows that EventNMF reduces to Poisson-NMF in the degree-0 case, while higher-degree splines avoid binning and yield smooth continuous-time factors. The proof is given in Appendix~\ref{proof:histogram-nmf}.

\begin{proposition}[Reduction to Poisson loss]\label{prop:histogram-nmf}
For a degree-0 basis, the negative log-likelihood satisfies
\begin{equation*}
\mathcal{L}(\{\mathbb{Y}_i\};\, U, \Gamma)
=\mathrm{const}+ \sum_{i,b} \bigl(\hat{X}_{ib} - X_{ib}\log \hat{X}_{ib}\bigr),
\end{equation*}
i.e., it reduces to the Poisson loss up to terms independent of the parameters.
\end{proposition}

\ptitle{Extension to dynamic networks.}
EventNMF extends naturally to dynamic networks, where observations are interaction events $\mathcal{E}=\{(s_m, d_m, t_m)\}_{m=1}^M$ between a source node $s_m$ and a destination node $d_m$ at time $t_m$,
with applications in social network analysis and cybersecurity~\cite{modellIntensityProfileProjection2023, romeroMultiresolutionAnalysisStatistical2025}.
To extend EventNMF to dynamic networks, we model this data as a collection of counting measures $\{\Ybb_{ij}\}_{1\leq i,j \leq N}$, each following a Poisson process with intensity
$\lambda_{ij}(t) = (u_i \odot v_j)^\top \Gamma\, \Phi(t)$,
where $u_i, v_j \in \mathbb{R}_+^R$ are source and target node loadings.
The fitting procedure mirrors the single-entity case; we provide the full gradients and multiplicative updates in Appendix~\ref{appendix:dynamic_networks}.
When a degree-0 basis is used, the model further reduces to non-negative tensor factorization~\cite{cichockiNovelMultilayerNonnegative2007a, gauvinDetectingCommunityStructure2014} with a KL divergence loss, rather than the Frobenius norm used in standard NTF.

%% file: figure1.tex
\begin{figure}[t]
\centering
%
\newlength{\topheight}
\newlength{\botheight}
\setlength{\topheight}{3.5cm}  
\setlength{\botheight}{4.2cm}  
%
\begin{minipage}[t]{0.49\textwidth}
\centering
\vbox to \topheight{
    \vfill
    \includegraphics[width=\textwidth]{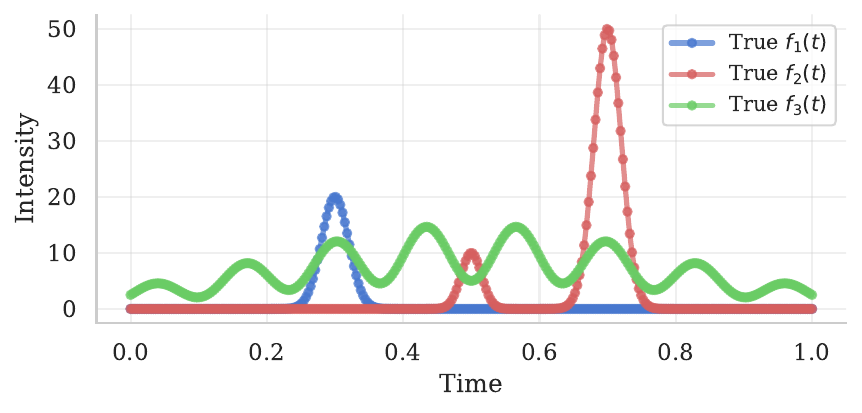}
    \vfill
}
\vspace{-0.5em}
\subcaption{True temporal factors.}
\label{fig:true_temporal_factors}
\vbox to \botheight{
    \vfill
    \includegraphics[width=\textwidth]{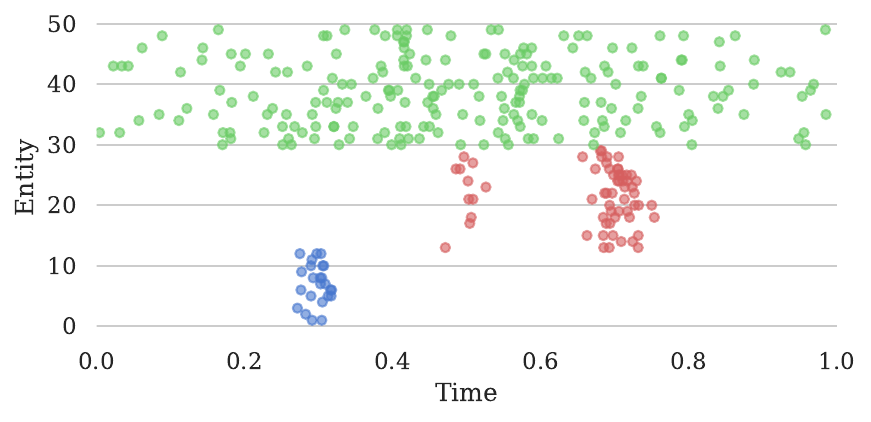}
    \vfill
}
\vspace{-0.5em}
\subcaption{Raster plot of event data.}
\label{fig:event_data_raster}
\end{minipage}
\hfill
\begin{minipage}[t]{0.49\textwidth}
\centering
\vbox to \topheight{
    \vfill
    \includegraphics[width=\textwidth]{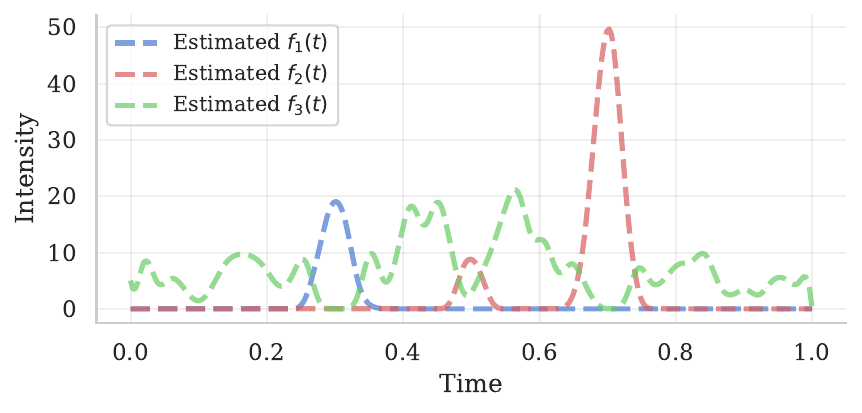}
    \vfill
}
\vspace{-0.5em}
\subcaption{Estimated temporal factors.}
\label{fig:estimated_temporal_factors}
\vbox to \botheight{
    \vfill
    \includegraphics[width=0.6\textwidth]{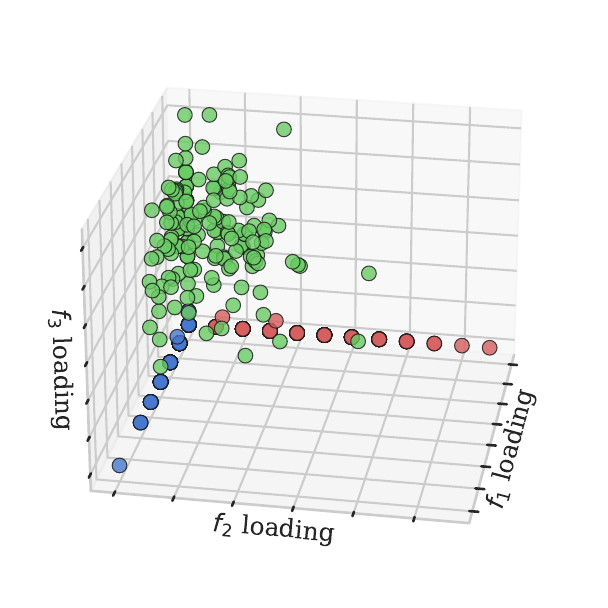}
    \vfill
}
\vspace{-0.5em}
\subcaption{Estimated factor loadings $U$.}
\label{fig:entity_embedding}
\end{minipage}
\caption{Illustration of our EventNMF method on a synthetic example with $R=3$ orthogonal factors. \textbf{Left column:} synthetic data:
true latent factors (top) and raster plot of observed events (bottom). \textbf{Right column:}
Output of EventNMF: recovered temporal factors (top) and entity factor loadings $U$ (bottom) (each point corresponds to an entity, i.e. a row in the raster plot (bottom left)). The colors of the points in the embedding correspond to the true cluster labels of the entities.}
\label{fig:example_2factors}
\vspace{-1.2em}

\end{figure}

\setlength{\topheight}{0pt}
\setlength{\botheight}{0pt}

%% file: experiments_synthetic.tex
\section{Synthetic Experiments\label{sec:simulations}}
To evaluate EventNMF, we consider a synthetic setting with known latent factors and assignment structure. This allows us to (i) assess recovery of latent factors and loadings, (ii) compare against a binned NMF baseline, and (iii) study bias-variance tradeoffs and the effect of the number of factors.

\subsection{Experimental setting}
We simulate $N$ entities over the time horizon $T=1$, partitioned into $R=3$ groups. Each entity follows an inhomogeneous Poisson process with intensity $\lambda_i(t)$ given by a nonnegative combination of $R=3$ latent temporal factors $f_r(t)$ with group-specific loadings $u_{ir}$. Each group corresponds to a distinct ground-truth factor: a Gaussian bump ($f_1$), a double-peaked function ($f_2$), and a sinusoidally modulated envelope ($f_3$), as illustrated in Figure~\ref{fig:true_temporal_factors}. Loadings are one-hot group indicators, yielding an exact rank-$3$ structure. Additional details are provided in Appendix~\ref{appendix:synthetic_data}.

\idea{
  \ptitle{Methods.}
  We compare two instantiations of the EventNMF objective:
  \begin{itemize}
    \item \textbf{EventNMF (Splines):} Our method with a B-spline basis of degree 4 and $K=40$ internal knots.
    \item \textbf{Binned NMF ($B$):} The same model with a histogram basis $\phi_b(t) = \ind_{I_b}(t)$, equivalent to applying standard Poisson NMF to binned event counts. We vary $B \in \{20, 40, 100\}$ to assess sensitivity to bin resolution.
    \item \textbf{pp-seq} \cite{}
  \end{itemize}
  Both methods use multiplicative updates (Algorithm~\ref{alg:eventnmf}) with $R=3$ factors and 200 iterations.

  For the identifiability experiments (Section~\ref{sec:identif}), we also use projected gradient descent with Adam optimization as an alternative to multiplicative updates.
}

\ptitle{Metrics.}
The quality of model fit is quantified using three complementary metrics. First, the \emph{negative log-likelihood} (NLL) on both train and held-out test events, where test events are obtained by Bernoulli thinning with $p_{\mathrm{train}}=0.8$, ensuring independence between the two sets. Second, since we have access to the true intensities $\lambda_i(t)$, we compute the \emph{normalized mean squared error} (NMSE) between the true and the estimated intensities $\hat{\lambda}_i(t)$, averaged across entities. This allows us to assess the accuracy of the entity-level intensity estimates. Finally, in order to evaluate the recovery of the synthetic latent factors, we compute the \emph{normalized factor integrated squared error} (NFISE) between true and recovered latent factors, after optimal permutation alignment (since the factors are identifiable only up to permutation). Formal definitions are given in Appendix~\ref{app:metrics}.

\subsection{Results}
\begin{figure}[h!]
  \centering
  \begin{subfigure}[b]{0.32\columnwidth}
    \includegraphics[width=\textwidth]{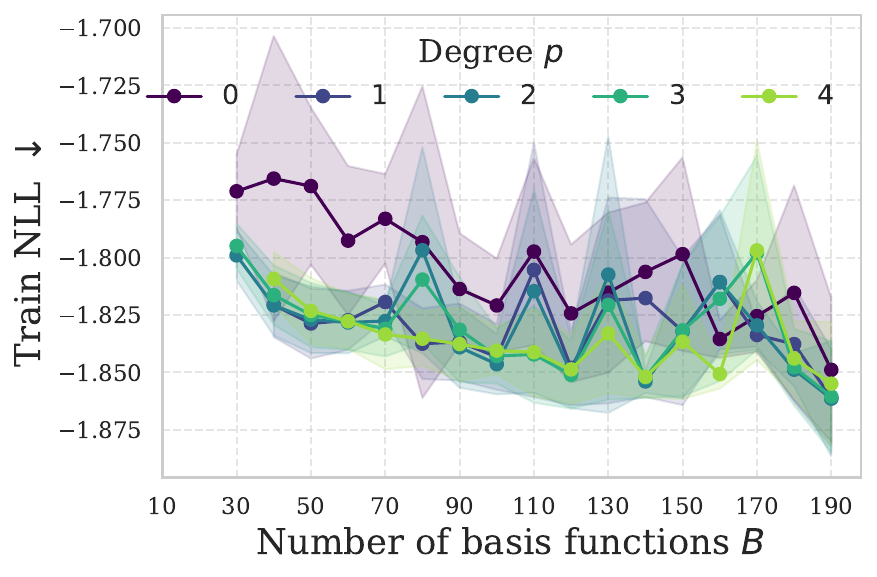}
    \caption{Train NLL}
    \label{fig:basis-sweep-train-nll}
  \end{subfigure}
  \hfill
  \begin{subfigure}[b]{0.32\columnwidth}
    \includegraphics[width=\textwidth]{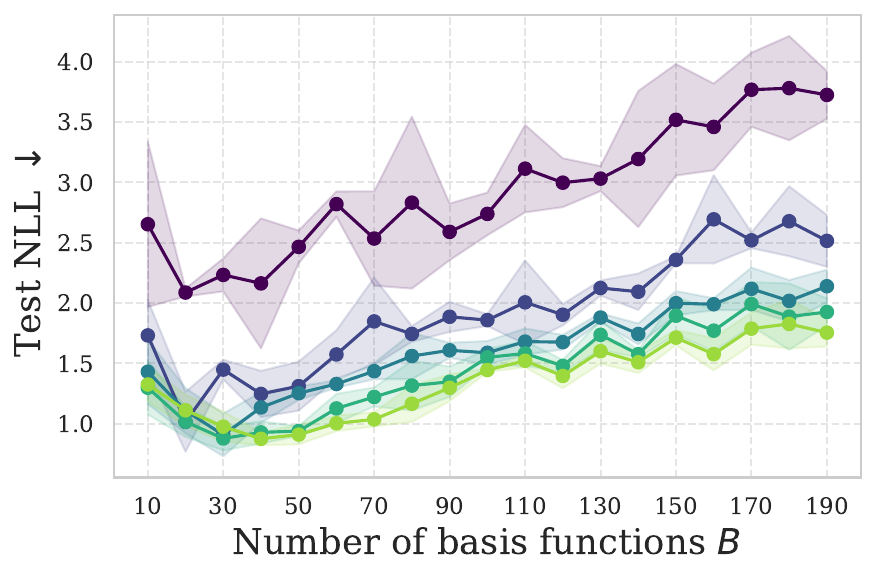}
    \caption{Test NLL}
    \label{fig:basis-sweep-test-nll}
  \end{subfigure}
  \hfill
  \begin{subfigure}[b]{0.32\columnwidth}
    \includegraphics[width=\textwidth]{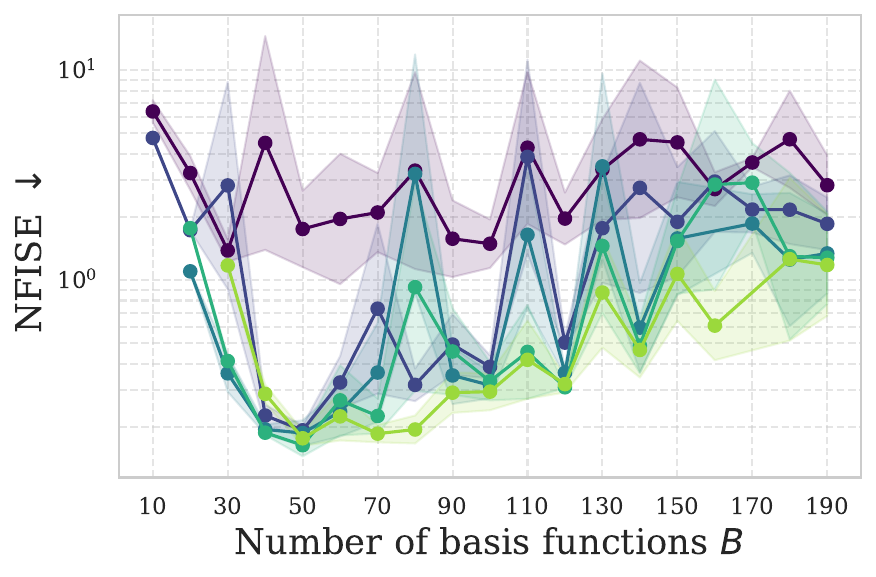}
    \caption{NFISE}
    \label{fig:basis-sweep-fise}
  \end{subfigure}
  \caption{Train NLL (left), Test NLL (centre), and NFISE (right) vs.\ number of basis functions $B$ for each spline degree $p \in \{0,\ldots,4\}$. Train/test split via Bernoulli thinning ($p_{\mathrm{train}}=0.8$). The shaded bands show $\pm 1$ std over 3 seeds.}
  \label{fig:basis-sweep}
\end{figure}

\ptitle{Effect of hyperparameters.} In order to study the bias-variance tradeoff in EventNMF, we sweep over spline degree $p \in \{0,\ldots,4\}$ and number of basis functions $B$. Figure~\ref{fig:basis-sweep-train-nll} shows that increasing $B$ leads
to a plateau in training NLL across all degrees, indicating overfitting.
Due to the smoothness of the ground-truth latent factors, $p=0$ lacks
sufficient expressivity and is outperformed by higher-degree bases.
Conversely, $p=4$ introduces more coefficients, which may render the
optimization unstable. On the test set (Figure~\ref{fig:basis-sweep-test-nll}), degrees $p=0$ and $p=1$ underfit the smooth latent processes, leading to worse performance than higher-degree spline bases. The optimal configuration is achieved at $p=3$ and
$B=30$, illustrating the bias-variance trade-off inherent to EventNMF.
This trade-off is further reflected in
Figure~\ref{fig:basis-sweep-fise}, where, for $p>1$, the NFISE stabilises
around $B=30$-$50$ before increasing again. The relative stability
across this range suggests robustness to the choice of $B$, as redundant
components in the spline basis are effectively driven toward zero.

\newcommand{\archive}[1]{}
\archive{
  Table~\ref{tab:synthetic_benchmark} shows that EventNMF (Splines) achieves the lowest NMSE ($27.78 \pm 0.00$) and the best NLL, outperforming all binned variants. The best binned baseline (B=100) achieves NMSE $= 30.36 \pm 3.61$, an 8.5\% increase over the spline model, reflecting the artificial discontinuities introduced by the histogram basis. Binned NMF also exhibits substantially higher variance across seeds, particularly at finer resolutions (B=40, 100), suggesting optimization instability from the piecewise-constant approximation. Loading recovery (MSE) is comparable across all methods.Figure~\ref{fig:factor-comparison} visualizes the learned factors for one replicate (after normalization to unit integral), confirming that the spline basis produces smooth, accurate reconstructions, while the histogram basis introduces staircase artifacts.
}

\archive{
  \ptitle{Q1: spline basis choice.}

  We first ask how the spline basis affects recovery. Figure~\ref{fig:basis-sweep} sweeps over $p \in \{0,1,2,3,4\}$ and $B \in \{5,10,20,50\}$, reporting train NLL, test NLL, and NMSE on the recovered latent factors after optimal Hungarian alignment. Degree $p=0$ is the histogram basis; higher degrees produce smoother intensity estimates. The main pattern is that increasing $B$ improves all metrics until the basis is expressive enough, after which performance plateaus. Moderate degrees $p \in \{1,2\}$ reach that plateau fastest on this synthetic data, while very high-degree splines need more knots to avoid oscillatory behaviour near the sharp peaks. We therefore use $p=3$ with $B=50$ as the default basis in the synthetic benchmark.}

\idea{
  things to say:
  - As seen on \ref{fig:basis-sweep-train-nll}, for all degrees, increasing $B$ reaches a plateau in terms of Train NLL, indicating overfitting. Due to the smoothness of ground truth latent factors, degree $p=0$ is not enough to reach a good nll, and is in general outperformed by higher degree bases.
  In contrast, $p=4$ implies more coefficients, which may be at the detriment of fitting stability. This explains that the train nll is not goes up with this p=4.

  - On Test NL \ref{fig:basis-sweep-test-nll}, we can see that p=0 and p=1 do not capture the underlying processes properly, as they lead to poor testNLL compared to higher degrees. We can see that the optimum is obtained for p=3 and B=30. This illustrates clearly the bias-variance tradeoff inherent to EventNMF.

  - The same bias-variance tradeoff is visible on Figure \ref{fig:basis-sweep-fise}, where it can be seen that after 30-50 bases, for degrees higher than 1, the NFISE plateaus, before increasing again. The fact that the NFISE stays more or less stable for a certain range of B values is a good sign of the robustness of the method to the choice of the number of basis functions, since it will automatically push to zeros.....
}

\input{experiment_sparsity.tex}

\archive{We next vary event density by scaling all intensity functions by $\alpha \in [0.05, 5.0]$, which changes the average number of events per entity from roughly $M/N \approx 0.3$ to $M/N \approx 27$. Figure~\ref{fig:sparsity-curve} compares EventNMF (splines) with binned NMF and reports NMSE and NLL as a function of this scaling. The gap is largest in the sparse regime: when events are scarce, EventNMF recovers smoother factors and lower likelihood error, while the histogram baseline degrades more quickly. As density increases, the two methods become closer because the binning approximation becomes less harmful. This is the regime where continuous-time modeling matters most.
}

\archive{
  \ptitle{Q6: Identifiability and factor recovery.}
  \label{sec:identif}

  A fundamental question for any matrix factorization method is: \emph{under what conditions can the latent factors be reliably recovered?}
  We answer this empirically by varying three axes independently, keeping all other settings fixed at $p=3$, $B=100$, $R^*=3$, and reporting two complementary metrics: NMSE on the recovered temporal factors (after optimal Hungarian permutation alignment) and the Adjusted Rand Index (ARI) of entity-loading cluster assignments against the true groups.
  Results are averaged over 10 random seeds; shaded bands show $\pm 1$ standard deviation.

  \begin{figure}[t]
    \centering
    \begin{subfigure}[b]{0.32\columnwidth}
      \includegraphics[width=\textwidth]{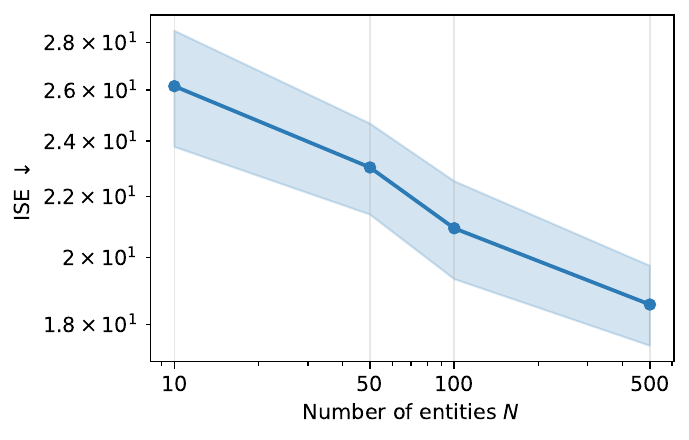}
      \caption{Factor NMSE}
    \end{subfigure}
    \hfill
    \begin{subfigure}[b]{0.32\columnwidth}
      \includegraphics[width=\textwidth]{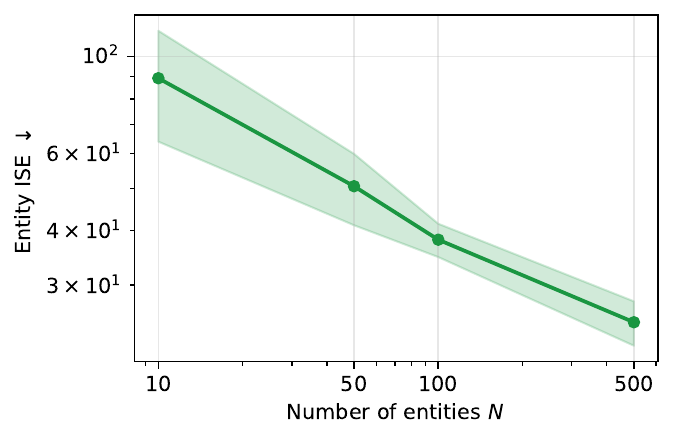}
      \caption{Residuals}
    \end{subfigure}
    \hfill
    \begin{subfigure}[b]{0.32\columnwidth}
      \includegraphics[width=\textwidth]{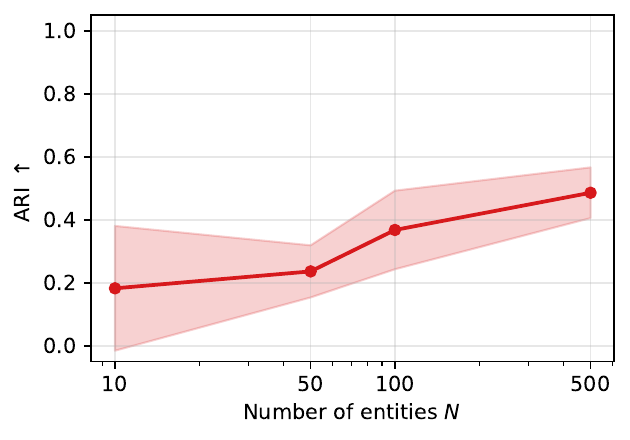}
      \caption{ARI}
    \end{subfigure}
    \caption{Factor NMSE, residuals, and ARI vs.\ number of entities $N$.
      Factor recovery reaches near-optimal performance at $N \approx 50$.
      Shaded bands: $\pm 1$ std over 10 seeds.}
    \label{fig:identif-N}
  \end{figure}

  \begin{figure}[t]
    \centering
    \begin{subfigure}[b]{0.32\columnwidth}
      \includegraphics[width=\textwidth]{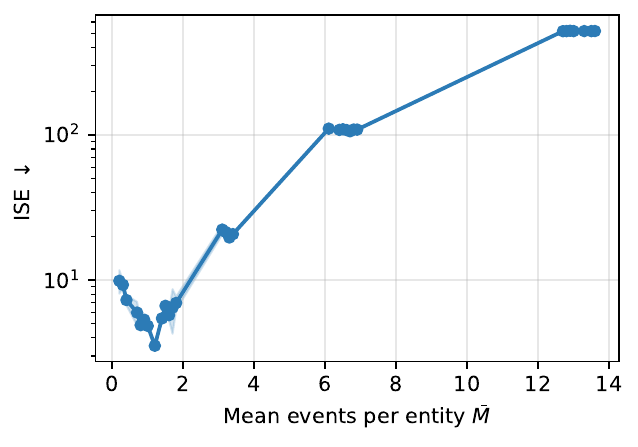}
      \caption{Factor NMSE}
    \end{subfigure}
    \hfill
    \begin{subfigure}[b]{0.32\columnwidth}
      \includegraphics[width=\textwidth]{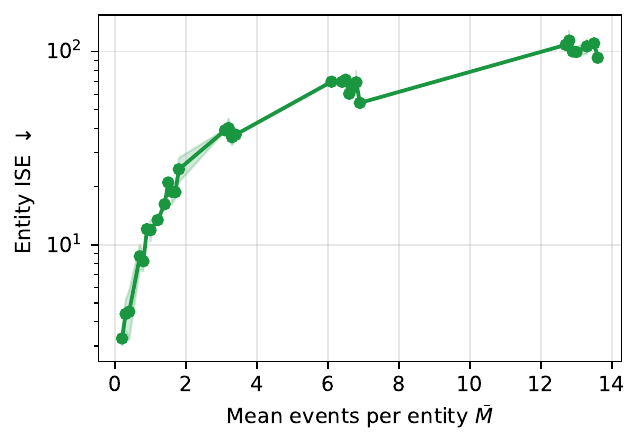}
      \caption{Residuals}
    \end{subfigure}
    \hfill
    \begin{subfigure}[b]{0.32\columnwidth}
      \includegraphics[width=\textwidth]{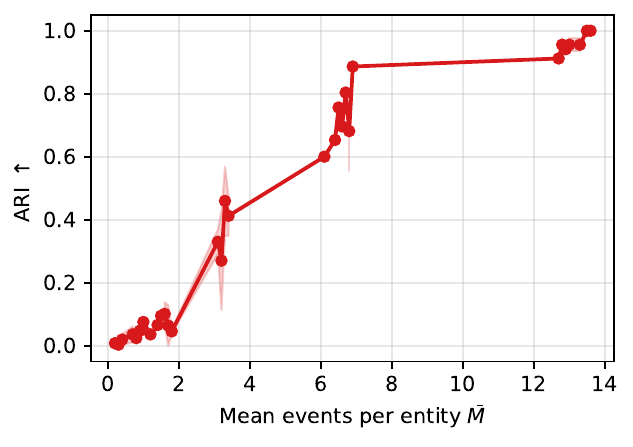}
      \caption{ARI}
    \end{subfigure}
    \caption{Factor NMSE, residuals, and ARI vs.\ mean events per entity $\bar{M}$.
      Recovery improves steeply in the sparse regime and plateaus beyond $\bar{M}\approx 40$.
      Shaded bands: $\pm 1$ std over 10 seeds.}
    \label{fig:identif-M}
  \end{figure}

  \begin{figure}[t]
    \centering
    \begin{subfigure}[b]{0.32\columnwidth}
      \includegraphics[width=\textwidth]{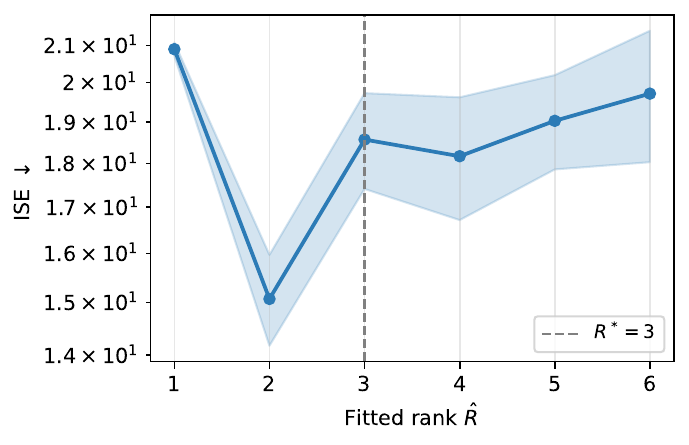}
      \caption{Factor NMSE}
    \end{subfigure}
    \hfill
    \begin{subfigure}[b]{0.32\columnwidth}
      \includegraphics[width=\textwidth]{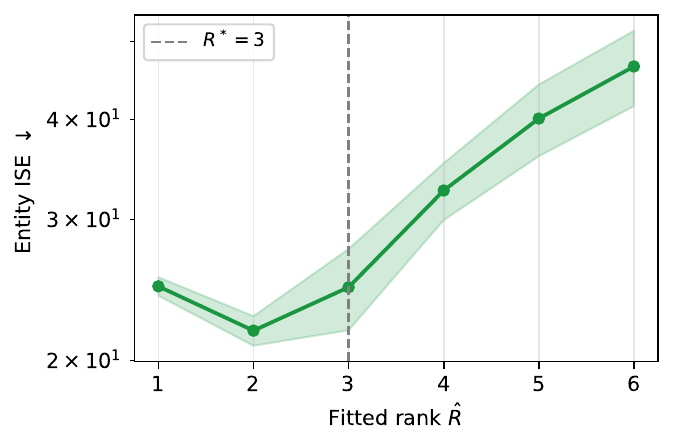}
      \caption{Residuals}
    \end{subfigure}
    \hfill
    \begin{subfigure}[b]{0.32\columnwidth}
      \includegraphics[width=\textwidth]{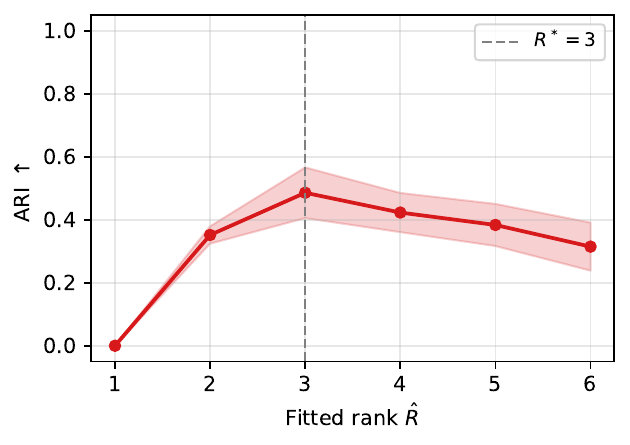}
      \caption{ARI}
    \end{subfigure}
    \caption{Factor NMSE, residuals, and ARI vs.\ fitted rank $\hat{R}$ (dashed: true rank $R^*=3$).
      Underfitting causes sharp degradation in all metrics; mild overfitting has little impact.
      Shaded bands: $\pm 1$ std over 10 seeds ($N=500$).}
    \label{fig:identif-Rhat}
  \end{figure}
}

\ptitle{Comparison with existing methods.} Despite event data being ubiquitous, relatively few methods target these
datasets directly. Most approaches rely on binning or smoothing to
convert event streams into standard temporal signals before applying
factorization.
We discuss the relationship of EventNMF to the two closest existing methods, NARFD and PPCA, and report empirical comparisons on the synthetic dataset.

\begin{figure}[t]
  \centering
  \begin{subfigure}[t]{0.32\linewidth}
    \centering
    \includegraphics[width=\linewidth]{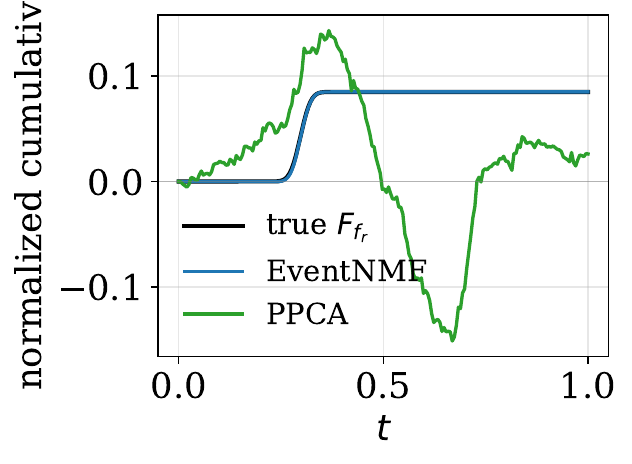}
    \caption{Factor 1}
  \end{subfigure}\hfill
  \begin{subfigure}[t]{0.32\linewidth}
    \centering
    \includegraphics[width=\linewidth]{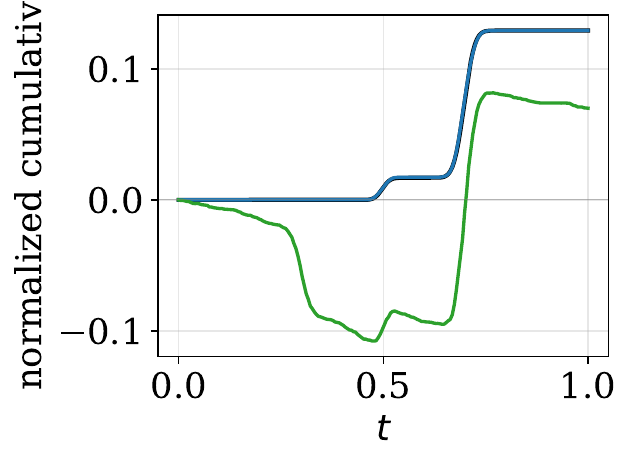}
    \caption{Factor 2}
  \end{subfigure}\hfill
  \begin{subfigure}[t]{0.32\linewidth}
    \centering
    \includegraphics[width=\linewidth]{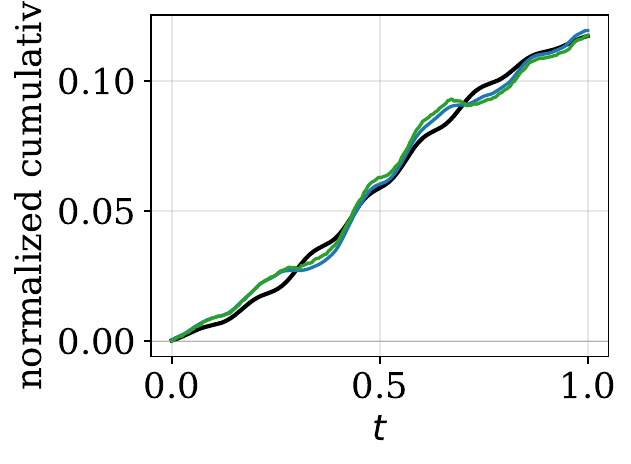}
    \caption{Factor 3}
  \end{subfigure}
  \caption{Comparison with PPCA in terms of recovered factor cumulatives. True (black), EventNMF (blue), PPCA (green), normalized to unit $L^2$ norm.}
  \label{fig:cumulative-comparison}
  \vspace{-1em}
\end{figure}

\ptitle{PPCA}~\cite{picardPCAPointProcesses2024} applies functional PCA to empirical cumulatives of the input counting measures $F_i(t) = \Ybb_i([0,t])$, yielding orthonormal signed eigenfunctions $\{F_{\mu_r}\}$ and real-valued scores $\xi_{i,r}$. Unlike EventNMF, PPCA yields signed scores and orthonormal components rather than nonnegative loadings and rate templates. This is a structural limitation: by~\citep[Thm.~4.4]{picardPCAPointProcesses2024}, the $r$-th eigenfunction must have $r-1$ sign changes, ruling out monotone cumulatives. As shown in Figure~\ref{fig:cumulative-comparison}, EventNMF tracks all true factor cumulatives accurately, while PPCA matches only the near-linear-trend factor.

\begin{wrapfigure}{r}{0.4\columnwidth} 
  \vspace{-1em} 
  \centering 
  \includegraphics[width=0.4\columnwidth]{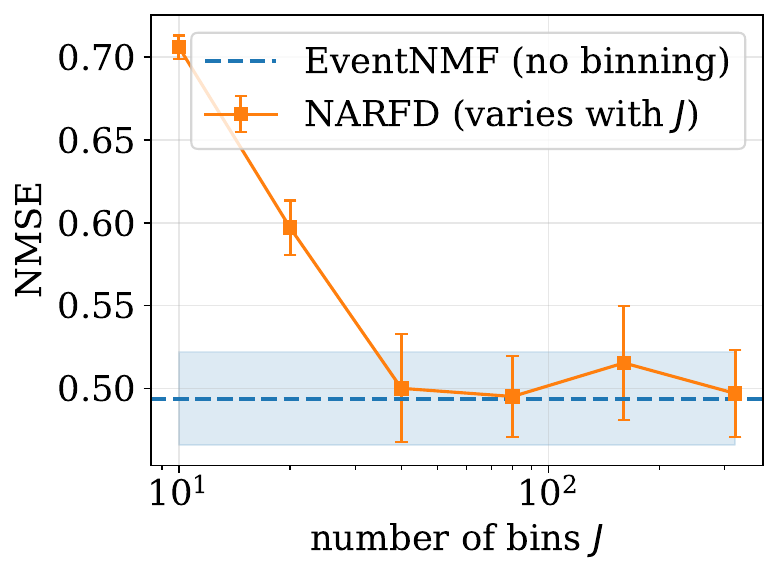} 
  \caption{Comparison with NARFD. The error bars and shaded band show $\pm 1$ std over 5 seeds.} \label{fig:bins-sensitivity} 
  \vspace{-1em} 
\end{wrapfigure}

\ptitle{NARFD}~\cite{backenrothNonnegativeDecompositionFunctional2020} models longitudinal count data as
$Y_i(t_j) \sim \text{Poisson}(\mu_{ij})$ with
$\mu_{ij} = u_i^\top \Gamma \Phi(t_j)$, where $\Phi(t)$ is a B-spline basis.
For event data, NARFD discretizes the time axis into $J$ bins, making the input observations depend on the choice of bin width. EventNMF can be viewed as a continuous-time counterpart of NARFD: both perform a nonnegative Poisson factorization over spline bases, but EventNMF operates directly on raw event times and therefore avoids binning. As shown in Figure~\ref{fig:bins-sensitivity}, NARFD performance depends strongly on the number of bins $J$: large bins oversmooth temporal structure, while small bins lead to sparse counts and increased variance. In contrast, EventNMF achieves comparable performance to NARFD without requiring a binning parameter.

%% file: experiment_sparsity.tex
\begin{wrapfigure}{r}{0.6\textwidth}
\vspace{-1em}
\centering
\begin{subfigure}[b]{0.48\linewidth}
\includegraphics[width=\textwidth]{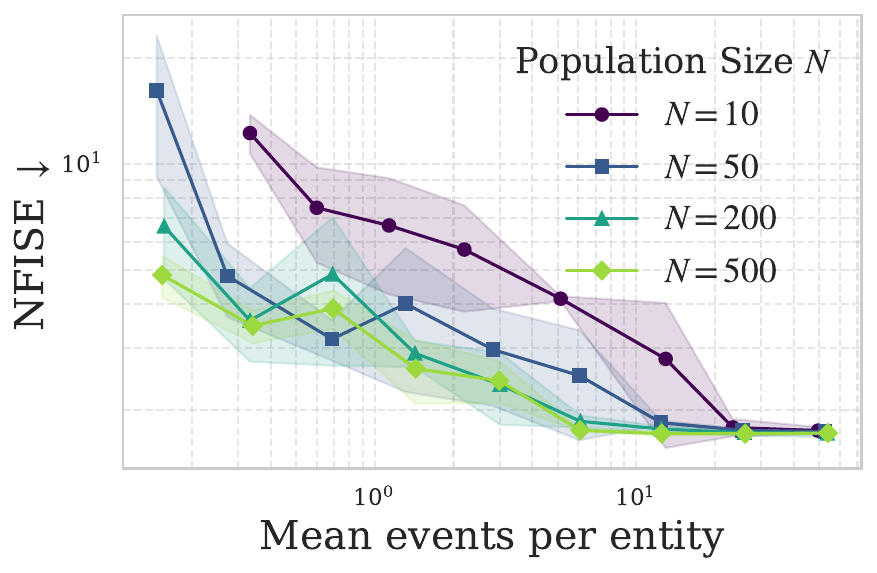}
\caption{NFISE vs.\ events per entity}
\label{fig:sparsity-nfise}
\end{subfigure}
\hfill
\begin{subfigure}[b]{0.48\linewidth}
\includegraphics[width=\textwidth]{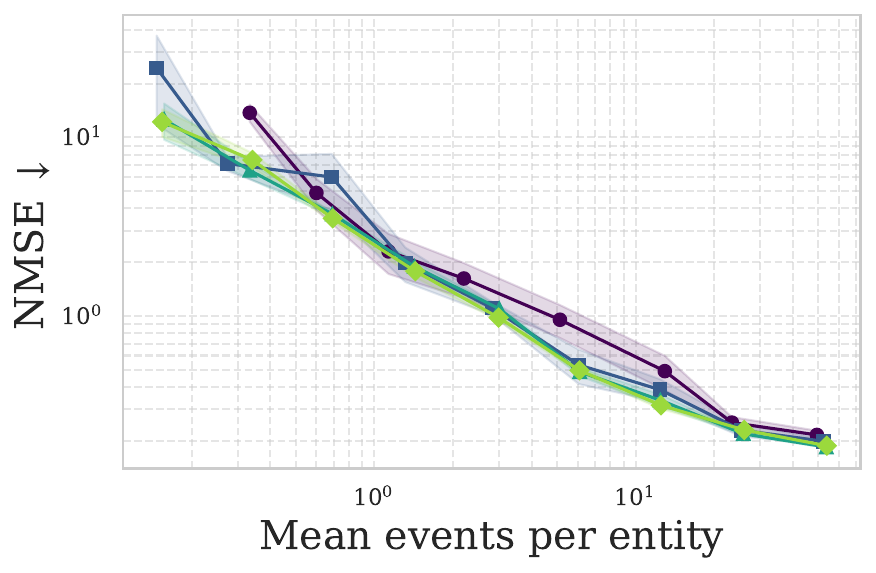}
\caption{NMSE vs.\ events per entity}
\label{fig:sparsity-mse}
\end{subfigure}
\caption{\textbf{Sparsity robustness across population size $N$ and event density.} (a): normalized NMSE between learned and true factor shapes, after permutation matching. (b): normalized MSE between true and learned entity-level intensities. Shaded bands show $\pm 1$ standard deviation over 3 seeds.}
\label{fig:sparsity-sweep}
\vspace{-1em}
\end{wrapfigure}
\ptitle{Effect of data sparsity.}
To study how EventNMF performs under data sparsity, we vary the average number of events per entity from roughly $0.05$ to $50$ by rescaling the ground-truth intensities as $\lambda_i^{\star}(t) = \frac{\alpha}{N}\, u_i^{\star\top} \mathbf{f}^{\star}(t)$, where $\alpha$ is a global scale parameter which modulates the number of events per entity. Results are shown in Figure~\ref{fig:sparsity-sweep}. In the \textbf{sparse regime} (fewer than $\sim$1 event per entity), having more entities helps considerably: at $0.3$ events per entity, both NFISE and NMSE for $N=500$ are roughly $4\times$ lower than for $N=10$. In the \textbf{dense regime} ($\gtrsim 5$ events per entity), each entity provides sufficient information to reliably recover its latent intensity, so performance becomes largely independent of $N$ and the curves collapse. The error at the final plateau on Fig ~\ref{fig:sparsity-nfise} is due to a basis approximation limit: even when each entity has many events, the spline representation cannot capture temporal structure finer than the knot spacing, which sets a lower bound on the achievable error.

%% file: applications.tex

\input{earthquakes.tex}

\input{allen_neuropixel.tex}

\input{primary_school.tex}

%% file: earthquakes.tex
\subsection{Earthquake Occurrence Patterns}

\begin{wrapfigure}{r}{0.6\columnwidth}
\vspace{-2em}
  \centering
  \begin{minipage}[t]{0.3\columnwidth}
    \includegraphics[width=\linewidth]{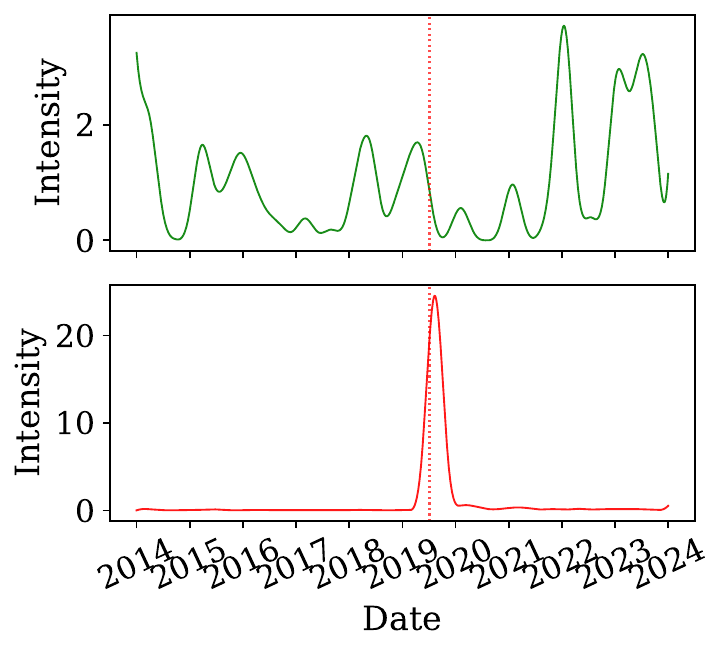}
  \end{minipage}%
  \hfill
  \begin{minipage}[t]{0.3\columnwidth}
    \includegraphics[width=\linewidth]{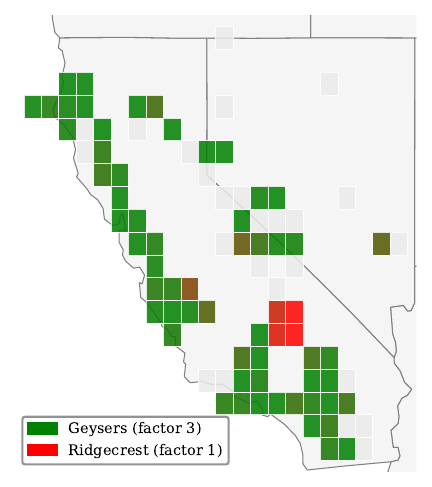}
  \end{minipage}
  \caption{Two factors recovered by EventNMF.
    \textbf{(a)}~Temporal intensity profiles of the Geysers factor
    (green) and the Ridgecrest factor (red); the dotted line marks
    the M7.1 earthquake (July 2019).
    \textbf{(b)}~Geographic map of cell loadings; each cell is coloured by a
    blend of its Geysers loading (green) and Ridgecrest loading (red--red).}
  \label{fig:eq-earthquake}
  \vspace{-1em}
\end{wrapfigure}

We apply EventNMF to ten years of seismic data (January 2014--January 2024) from the
USGS earthquake catalog for the California--Nevada region (magnitude $\geq 2.5$,
latitudes $33$--$42^\circ$N, longitudes $125$--$114^\circ$W; $14{,}612$ events total).
Space is discretised into $0.5^\circ$ grid cells ($N=88$ cells with $\geq 20$ events),
timestamps are normalised to $[0,1]$, and we fit EventNMF with $R=8$ components.

EventNMF separates the earthquake event data into interpretable factors, two of which are shown in Figure~\ref{fig:eq-earthquake}. 
One factor (red) is concentrated in a region of Southern California and has a sharp
temporal peak in July 2019, coinciding with the M7.1 Ridgecrest earthquake, followed by
a rapid decay a pattern typical of a mainshock and its aftershock sequence.
A second factor (green) is localised near The Geysers area in Northern California and
shows a much smoother temporal profile over the full ten-year window, consistent with a
persistent, slowly varying source of low-intensity activity.
The geographic map (\cref{fig:eq-earthquake}b) confirms that the two factors load on
distinct, non-overlapping regions, with little mixing between them.

%% file: allen_neuropixel.tex
\subsection{Single Trial Analysis on the Allen Neuropixels Dataset}
\begin{figure}[t]
  \centering
  \begin{minipage}[c]{0.38\textwidth}
    \begin{subfigure}[t]{\linewidth}
      \centering
      \includegraphics[width=\linewidth]{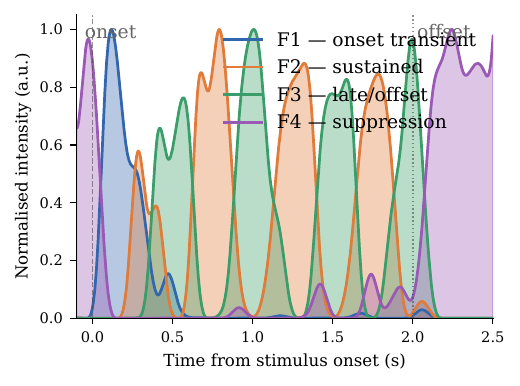}
      \caption{Learned temporal factors $f_r(t)$: onset transient (F1), sustained (F2),
        late/offset (F3), and suppression (F4).}
      \label{fig:allen-factors}
    \end{subfigure}

    \vspace{1em}

    \begin{subfigure}[t]{\linewidth}
      \centering
      \includegraphics[width=\linewidth]{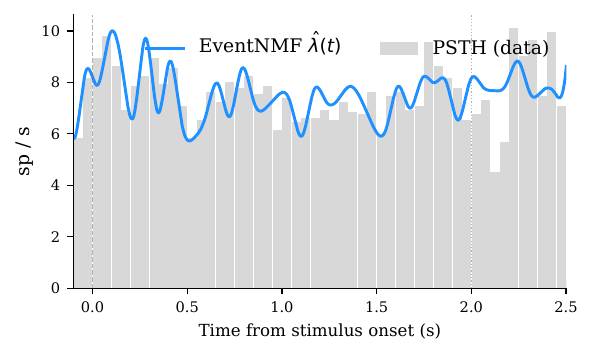}
      \caption{Population-mean reconstruction $\hat{\lambda}(t)$ vs.\ empirical PSTH
        for LP thalamic neurons.}
      \label{fig:allen-recon}
    \end{subfigure}
  \end{minipage}
  \hfill
  \begin{minipage}[c]{0.58\textwidth}
    \begin{subfigure}[t]{\linewidth}
      \centering
      \includegraphics[width=\linewidth]{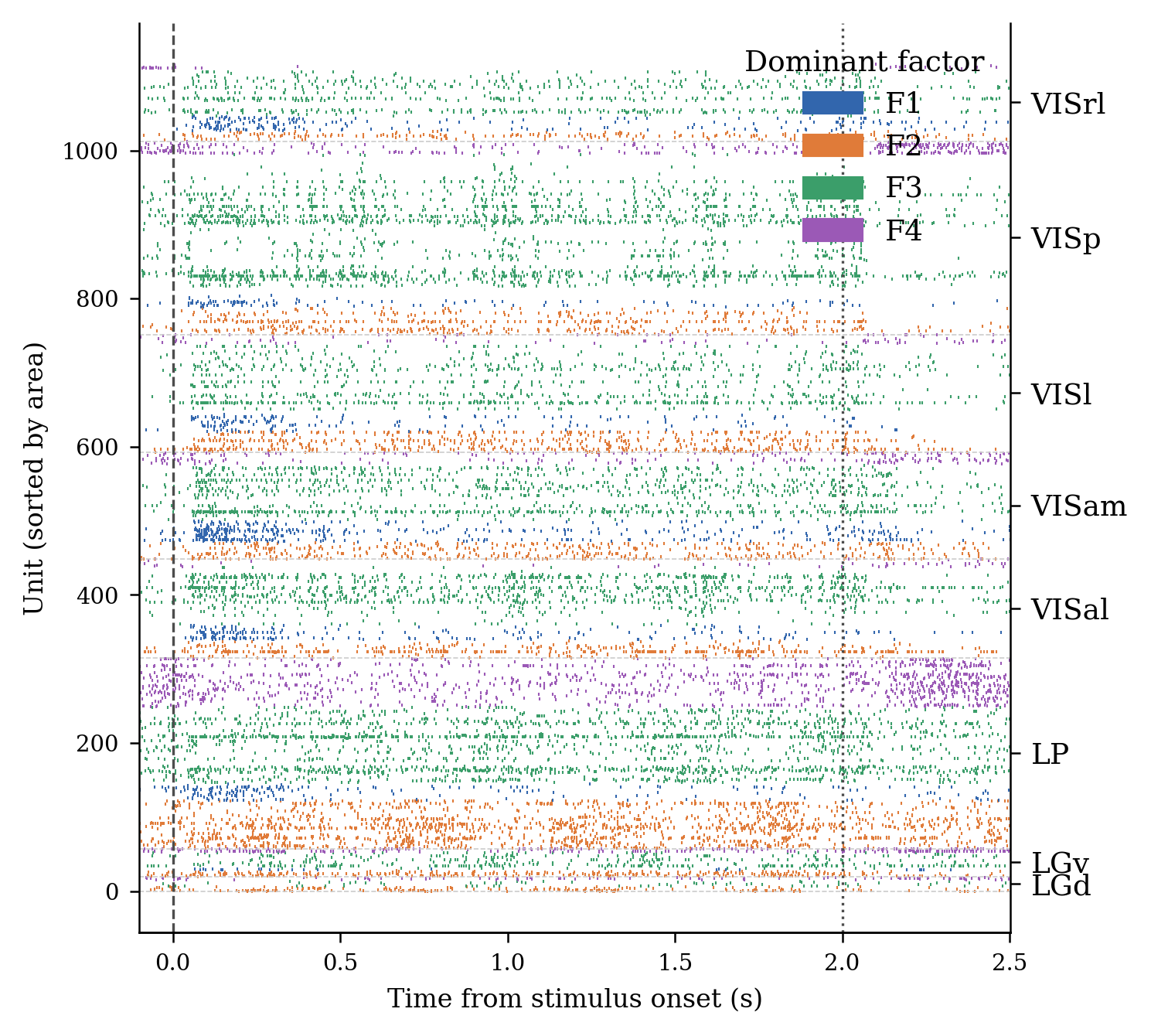}
      \caption{Spike raster sorted by brain area and coloured by dominant factor,
        revealing systematic area-specific temporal organisation.}
      \label{fig:allen-raster}
    \end{subfigure}
  \end{minipage}

  \caption{EventNMF analysis of a single trial from the Allen Visual Coding Neuropixels
    dataset. \textbf{(a)}~Four learned temporal factors capturing distinct population-level
    response motifs. \textbf{(b)}~Reconstruction vs.\ PSTH for LP thalamic neurons.
    \textbf{(c)}~Spike raster; dot colour indicates each unit's dominant factor,
    revealing systematic area-specific temporal organisation.}
  \label{fig:allen-summary}
\end{figure}

We apply EventNMF to the Allen Visual Coding Neuropixels dataset\footnote{\url{https://portal.brain-map.org/explore/circuits/visual-coding-neuropixels}}, a public recording of simultaneous spike trains from hundreds of neurons in a mouse viewing visual stimuli.
We restrict to visual and thalamic areas (VISp, VISl, VISal, VISam, VISrl, VISpm, LGd, LGv, LP) and fit EventNMF with $p=3, B=50, R=4$ on a single trial (90\textdegree\ orientation, 2\,Hz temporal frequency). As shown in Fig.~\ref{fig:allen-factors}, EventNMF recovers four temporally distinct factors: a sharp onset peak (F1), a rhythmically modulated sustained component (F2), a delayed component (F3), and a below-baseline component with a post-stimulus rebound (F4).
We then examine how factor assignments are distributed across brain areas.
Fig.~\ref{fig:allen-raster} shows the spike raster sorted by area, with each neuron coloured by its dominant factor. This allows us to visually identify which brain areas are most strongly associated with each factor.
Finally, Fig.~\ref{fig:allen-recon} shows that the population-mean reconstruction closely matches the empirical firing rate histogram for LP neurons, confirming that $R=4$ factors are sufficient to capture the main temporal structure in the data.

%% file: primary_school.tex
\subsection{Primary School Contact Network}
\begin{figure*}[t]
  \centering
  \begin{subfigure}[t]{0.4\linewidth}
    \centering
    \includegraphics[width=\linewidth]{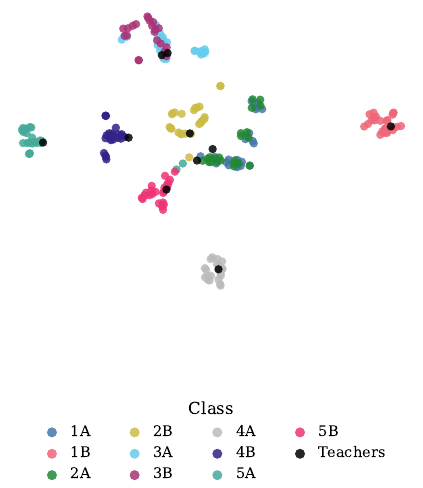}
    \caption{t-SNE of embeddings $W_i$, coloured by class.}
    \label{fig:ps_embeddings}
  \end{subfigure}
  \hfill
  \begin{subfigure}[t]{0.45\linewidth}
    \centering
    \includegraphics[width=\linewidth]{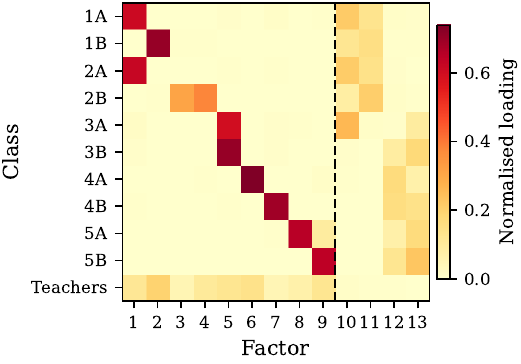}
    \caption{Class--factor affinity.}
    \label{fig:ps_affinity}
  \end{subfigure}
  \vspace{0.4em}
  \begin{subfigure}[b]{\linewidth}
    \centering
    \includegraphics[width=\linewidth]{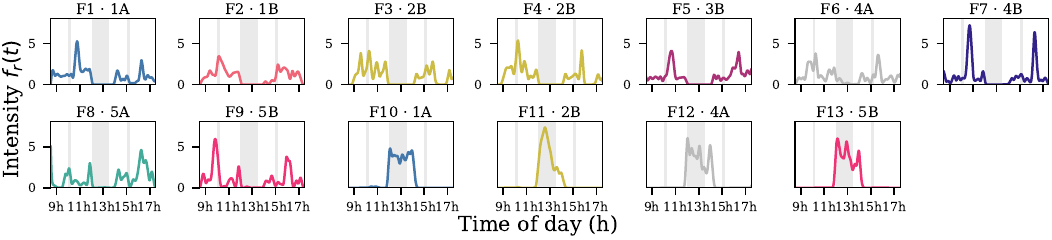}
    \caption{Factor profiles $f_r(t)$: flat $=$ within-class lessons;
             peaked $=$ free-mixing periods. Color and labels indicate dominant class for each factor.}
    \label{fig:ps_factors}
  \end{subfigure}
  \caption{EventNMF on the primary school dataset ($R = 13$, $p=3$, $B = 50$ knots)
           recovers all class groups unsupervised and distinguishes lesson-time
           from free-period interaction.}
  \label{fig:primary_school}
\end{figure*}

We apply EventNMF to the SocioPatterns primary school dataset~\citep{stehleHighResolutionMeasurementsFacetoFace2011a}, which records face-to-face contacts among 242 individuals across 11 classes over one school day. Following \cite{gauvinDetectingCommunityStructure2014}, we fit $R=13$ factors using a cubic spline basis with $B=50$ knots. Results are shown in \Cref{fig:primary_school}. The class-factor affinity matrix (\Cref{fig:ps_affinity}) shows that most factors correspond closely to a single ground-truth class without supervision. Older classes (4A--5B) produce nearly pure factors, whereas younger classes (1A--3B) exhibit more diffuse loadings, suggesting more interactions between classes. Teachers do not form a dedicated factor; instead, their loadings are spread across multiple factors, reflecting their movement between classrooms. The temporal factors (\Cref{fig:ps_factors}) reveal two distinct patterns: steady within-class interactions during lesson time and sharp peaks during lunch and recess when students from different classes interact more freely. Younger and older classes peak at slightly different times, consistent with staggered schedules.

%% file: appendix.tex
\clearpage

\section{Full Algorithms}
A first approach for fitting the EventNMF model is to use multiplicative updates, which are a popular choice for NMF problems due to their simplicity and guaranteed non-negativity. The update rules for $U$ and $\Gamma$ can be derived from the gradients of the loss function with respect to these parameters, leading the multiplicative update scheme as described in Algorithm \ref{alg:eventnmf}.

Another approach for optimization is HALS (hierarchical alternating least squares), where we update one coordinate at a time while keeping the others fixed. A convenient projected coordinate-descent form is
\begin{align*}
  u_{ir}      & \leftarrow \left[u_{ir} - \eta_{ir}\,\frac{\partial \Lcal}{\partial u_{ir}}\right]_+,           \\
  \gamma_{rb} & \leftarrow \left[\gamma_{rb} - \eta_{rb}\,\frac{\partial \Lcal}{\partial \gamma_{rb}}\right]_+,
\end{align*}
where $[x]_+ = \max(x,0)$ and $\eta_{ir},\eta_{rb}>0$ are coordinate-wise step sizes. In practice, this HALS-style update alternates over the rows of $U$ and the coefficients of $\Gamma$, enforcing non-negativity after each coordinate update.

Another approach for optimization is to use ALS, where we alternatively optimize with respect to $U$ and $\Gamma$ while keeping the other fixed, and set the negative coefficients to zero after each update to ensure non-negativity.
\begin{algorithm}[htbp]
  \caption{Multiplicative updates for EventNMF}
  \label{alg:eventnmf}
  \begin{algorithmic}[1]
    \REQUIRE Event sets $\{\Ecal_i\}_{i=1}^N$, basis functions $\{\phi_b\}_{b=1}^B$, rank $R$, number of iterations $K$
    \ENSURE Nonnegative factors $U\in\Rbb_+^{N\times R}$ and $\Gamma\in\Rbb_+^{R\times B}$
    \STATE Initialize $U$ and $\Gamma$ with positive entries
    \FOR{$k=1,\dots,K$}
    \FOR{$i=1,\dots,N$}
    \STATE Compute $\lambda_i(\tau)=u_i^\top \Gamma \Phi(\tau)$ for all $\tau\in\Ecal_i$
    \FOR{$r=1,\dots,R$}
    \STATE Update
    \[
      u_{ir}\leftarrow u_{ir}
      \frac{\sum_{\tau\in\Ecal_i}\frac{\sum_{b=1}^B \gamma_{rb}\phi_b(\tau)}{\lambda_i(\tau)}}
      {\sum_{b=1}^B \gamma_{rb} I(\phi_b)}
    \]
    \ENDFOR
    \ENDFOR
    \FOR{$r=1,\dots,R$}
    \FOR{$b=1,\dots,B$}
    \STATE Update
    \[
      \gamma_{rb}\leftarrow \gamma_{rb}
      \frac{\sum_{i=1}^N u_{ir}\sum_{\tau\in\Ecal_i}\frac{\phi_b(\tau)}{\lambda_i(\tau)}}
      {\sum_{i=1}^N u_{ir} I(\phi_b)}
    \]
    \ENDFOR
    \ENDFOR
    \ENDFOR
  \end{algorithmic}
\end{algorithm}

\begin{algorithm}[htbp]
  \caption{Alternating Least Squares for EventNMF}
  \label{alg:als-eventnmf}
  \begin{algorithmic}[1]
    \REQUIRE Event sets $\{\Ecal_i\}_{i=1}^N$, basis functions $\{\phi_b\}_{b=1}^B$, rank $R$, iterations $K$
    \ENSURE Nonnegative factors $U\in\Rbb_+^{N\times R}$ and $\Gamma\in\Rbb_+^{R\times B}$
    \STATE Initialize $U$ and $\Gamma$ with positive entries
    \FOR{$k=1,\dots,K$}
    \FOR{$i=1,\dots,N$}
    \STATE Fix $\Gamma$ and update $u_i$ by projected gradient or coordinate descent
    \STATE $u_i \leftarrow [\,u_i - \eta_U \nabla_{u_i} l_i(U,\Gamma)\,]_+$
    \ENDFOR
    \STATE Fix $U$ and update $\Gamma$ by projected gradient or coordinate descent
    \STATE $\Gamma \leftarrow [\,\Gamma - \eta_\Gamma \nabla_{\Gamma} \mathcal{L}(U,\Gamma)\,]_+$
    \ENDFOR
  \end{algorithmic}
\end{algorithm}

\clearpage
\section{Technical Proofs}

\subsection{Derivation of Gradients with Respect to $U$ and $\Gamma$\label{appendix:proof_gradients} (Proposition \ref{prop:gradients})}

Here we derive the gradient expressions stated as propositions in the method section, and show how they lead to the multiplicative update rules in Algorithm~\ref{alg:eventnmf}.

\ptitle{Setup.}
Recall from the method section that the latent factors are expanded in the basis $\{\phi_b\}_{b=1}^B$ as $f_r(t) = \sum_{b=1}^B \gamma_{rb}\,\phi_b(t)$, giving the compact intensity $\lambda_i(t) = u_i^\top \Gamma\,\Phi(t)$. The per-entity loss is
\begin{align*}
  l_i(U,\Gamma) = u_i^\top \Gamma\, I(\Phi) - \sum_{\tau \in \Ecal_i} \log\!\bigl(u_i^\top \Gamma\,\Phi(\tau)\bigr),
\end{align*}
where $I(\Phi) = \int_{\Tcal} \Phi(t)\,dt \in \Rbb^B$ is the vector of basis integrals. Throughout this section we use the shorthand $\lambda_i(\tau) \triangleq u_i^\top \Gamma\,\Phi(\tau) > 0$ for the fitted intensity at event time $\tau$, and split $l_i = A_i - B_i$ with
\begin{align*}
  A_i &= u_i^\top \Gamma\, I(\Phi) \\
  B_i &= \sum_{\tau \in \Ecal_i} \log u_i^\top \Gamma\,\Phi(\tau).
\end{align*}
The total loss is $\Lcal(U,\Gamma) = \sum_{i=1}^N l_i(U,\Gamma)$.

\subsubsection{Gradient with Respect to $u_{ir}$}

Differentiating $A_i$ with respect to the scalar $u_{ir}$:
\begin{align*}
  \frac{\partial A_i}{\partial u_{ir}} = \sum_{b=1}^B \gamma_{rb}\, I(\phi_b) = \bigl(\Gamma\, I(\Phi)\bigr)_r.
\end{align*}
Differentiating $B_i$ and applying the chain rule $\frac{\partial}{\partial u_{ir}}\log\lambda_i(\tau) = \frac{1}{\lambda_i(\tau)}\frac{\partial\lambda_i(\tau)}{\partial u_{ir}}$ we get:
\begin{align*}
  \frac{\partial B_i}{\partial u_{ir}}
  = \sum_{\tau \in \Ecal_i} \frac{\sum_{b=1}^B \gamma_{rb}\phi_b(\tau)}{\lambda_i(\tau)}
  = \sum_{\tau \in \Ecal_i} \frac{\bigl(\Gamma\,\Phi(\tau)\bigr)_r}{\lambda_i(\tau)}.
\end{align*}
Hence
\begin{align*}
  \frac{\partial l_i(U,\Gamma)}{\partial u_{ir}}
  = \bigl(\Gamma\, I(\Phi)\bigr)_r - \sum_{\tau \in \Ecal_i} \frac{\bigl(\Gamma\,\Phi(\tau)\bigr)_r}{\lambda_i(\tau)}.
\end{align*}
Stacking over $r = 1, \ldots, R$ gives us the vector gradient stated in the method:
\begin{align*}
  \nabla_{u_i} l_i(U,\Gamma) = \Gamma\, I(\Phi) - \sum_{\tau \in \Ecal_i} \frac{\Gamma\,\Phi(\tau)}{\lambda_i(\tau)}.
\end{align*}

\subsubsection{Gradient with Respect to $\gamma_{rb}$}

Differentiating $A_i$ with respect to the scalar $\gamma_{rb}$ gives:
\begin{align*}
  \frac{\partial A_i}{\partial \gamma_{rb}} = u_{ir}\, I(\phi_b).
\end{align*}
Differentiating $B_i$ via the chain rule yields:
\begin{align*}
  \frac{\partial B_i}{\partial \gamma_{rb}}
  = \sum_{\tau \in \Ecal_i} \frac{u_{ir}\phi_b(\tau)}{\lambda_i(\tau)}.
\end{align*}
Hence
\begin{align*}
  \frac{\partial l_i(U,\Gamma)}{\partial \gamma_{rb}}
  = u_{ir}\, I(\phi_b) - \sum_{\tau \in \Ecal_i} \frac{u_{ir}\phi_b(\tau)}{\lambda_i(\tau)}.
\end{align*}
Assembling over $r$ and $b$ via the outer product $(u_i \otimes v)_{rb} = u_{ir} v_b$ gives the matrix gradient for the loss associated with entity $i$:
\begin{align*}
  \nabla_{\Gamma} l_i(U,\Gamma) = u_i \otimes I(\Phi) - \sum_{\tau \in \Ecal_i} \frac{u_i \otimes \Phi(\tau)}{\lambda_i(\tau)}.
\end{align*}
Summing over entities gives $\nabla_{\Gamma} \Lcal(U,\Gamma) = \sum_{i=1}^N \nabla_\Gamma l_i(U,\Gamma)$, matching the proposition in the method section.

\subsubsection{Derivation of Multiplicative Updates}

Each scalar gradient decomposes as $\frac{\partial l_i}{\partial u_{ir}} = \gpU_{ir} - \gmU_{ir}$ and $\frac{\partial \Lcal}{\partial \gamma_{rb}} = \gpG_{rb} - \gmG_{rb}$, where the nonnegative positive and negative parts are
\begin{align*}
  \gpU_{ir} &= \sum_{b=1}^B \gamma_{rb}\, I(\phi_b), &
  \gmU_{ir} &= \sum_{\tau \in \Ecal_i} \frac{\sum_{b=1}^B \gamma_{rb}\phi_b(\tau)}{\lambda_i(\tau)}, \\[4pt]
  \gpG_{rb} &= \sum_{i=1}^N u_{ir}\, I(\phi_b), &
  \gmG_{rb} &= \sum_{i=1}^N u_{ir} \sum_{\tau \in \Ecal_i} \frac{\phi_b(\tau)}{\lambda_i(\tau)}.
\end{align*}
All four quantities are nonnegative since $u_{ir}, \gamma_{rb}, \phi_b \geq 0$ and $I(\phi_b) \geq 0$.

The multiplicative updates are deduced by choosing the coordinate-wise step sizes $\eta_{ir} = u_{ir}/\gpU_{ir}$ and $\eta_{rb} = \gamma_{rb}/\gpG_{rb}$ in the projected gradient scheme, which yields the update rules in Algorithm~\ref{alg:eventnmf} and preserves non-negativity of the iterates.

\ptitle{From gradient descent to multiplicative updates.}
A projected gradient step reads
\begin{align*}
  u_{ir} &\leftarrow \bigl[u_{ir} - \eta_{ir}\,(\gpU_{ir} - \gmU_{ir})\bigr]_+, &
  \gamma_{rb} &\leftarrow \bigl[\gamma_{rb} - \eta_{rb}\,(\gpG_{rb} - \gmG_{rb})\bigr]_+.
\end{align*}
Choosing the parameter-specific step sizes $\eta_{ir} = u_{ir}/\gpU_{ir}$ and $\eta_{rb} = \gamma_{rb}/\gpG_{rb}$ (positive as long as the iterates are positive), the update simplifies to
\begin{align*}
  u_{ir} &\leftarrow u_{ir}\,\frac{\gmU_{ir}}{\gpU_{ir}}, &
  \gamma_{rb} &\leftarrow \gamma_{rb}\,\frac{\gmG_{rb}}{\gpG_{rb}},
\end{align*}
and the projection $[\,\cdot\,]_+$ is automatically satisfied since both ratios are nonnegative. These are exactly the rules in Algorithm~\ref{alg:eventnmf}. Positivity of $u_{ir}$ and $\gamma_{rb}$ is preserved at every step because they enter only as positive scale factors.

\subsection{Proof of Proposition~\ref{prop:histogram-nmf}}
\label{proof:histogram-nmf}

Since $\Phi(t) = e_b$ for $t \in [k_b, k_{b+1})$, the intensity of entity $i$ is constant on each bin with value $u_i^\top \Gamma e_b$, giving expected count $\hat{X}_{ib} = u_i^\top \Gamma e_b\,(k_{b+1} - k_b)$. The negative log-likelihood for entity $i$ decomposes over bins as
\begin{align*}\ell_i(U, \Gamma)
&= \sum_{b=1}^{B} \Bigl[
    u_i^\top \Gamma e_b\,(k_{b+1} - k_b)
    - \mathbb{Y}_i([k_b, k_{b+1})) \log(u_i^\top \Gamma e_b)
\Bigr] \\
&= \sum_{b=1}^{B} \Bigl[
    \hat{X}_{ib} - X_{ib} \log \hat{X}_{ib}
\Bigr] + \mathrm{const},
\end{align*}
where the constant collects terms independent of $(U,\Gamma)$. Summing over $i$ yields
\begin{align*}
\mathcal{L}(\{\mathbb{Y}_i\};\, U, \Gamma)
= \mathrm{const} + \sum_{i=1}^N \sum_{b=1}^B \Bigl(
    \hat{X}_{ib} - X_{ib} \log \hat{X}_{ib}
\Bigr),
\end{align*}
which corresponds to the Poisson loss on the binned count matrix $X$. \qed

\subsection{Interpretation as a KL Divergence--NMF}
It is interesting to note that our framework is a particular instance of KL-divergence non-negative matrix factorization (NMF) studied in \cite{leeAlgorithmsNonnegativeMatrix}. Indeed, the population of counting measures $\{\Ybb_i\}_{i=1}^N$ defines an empirical distribution $\hat{\Pbb} = \frac{1}{N}\sum_{i=1}^N \delta_{\Ybb_i}$ supported on the space of counting measures, and for a given set of parameters $U, \Gamma$, the Poisson process model defines a distribution $\Pbb(\,\cdot\,;U,\Gamma)$ over the same space. It is well known that maximum likelihood estimation is equivalent to minimizing the KL divergence between the empirical distribution and the model \citep{bishopPatternRecognitionMachine2006, coverElementsInformationTheory2001}: the average negative log-likelihood decomposes as
\newcommand{\KL}{\mathrm{KL}}

\begin{equation}
    \frac{1}{N}\sum_{i=1}^N \ell(\Ybb_i; U, \Gamma) = \KL\!\left(\hat{\Pbb} \,\|\, \Pbb(\,\cdot\,;U,\Gamma)\right) + H(\hat{\Pbb}),
\end{equation}
where $H(\hat{\Pbb})$ is the entropy of the empirical distribution, which does not depend on the model parameters. Hence, minimizing the negative log-likelihood over $U, \Gamma$ is equivalent to minimizing $\KL(\hat{\Pbb} \| \Pbb(\,\cdot\,;U,\Gamma))$ with respect to the model parameters, placing our framework within the classical family of KL-divergence NMF methods.

\section{Extension to Dynamic Networks}
In the case of dynamic networks we have for each pair of entities $(i,j)$ a counting measure $\Ybb_{ij}$ and a corresponding unknown intensity function modeled as $\lambda_{ij}(t) = u_i \odot v_j \cdot \Gamma \Phi(t)$,
where $u_i \in \Rbb_+^R$ and $v_j \in \Rbb_+^R$ are the factor loadings for entities $i$ and $j$, $\Gamma \in \Rbb_+^{R\times B}$ is the matrix of basis coefficients, and $\Phi(t) = (\phi_1(t), \ldots, \phi_B(t))^T$ is the vector of basis functions.
The likelihood of the observed data is given by

\begin{align*}
  \Lcal(\{\Ybb_{ij}\}; U, V, \Gamma)
  = \sum_{i,j=1}^N l_{ij}(u_i, v_j, \Gamma), \quad
  l_{ij} = \int \lambda_{ij}(t)\,dt - \sum_{\tau \in \Ecal_{ij}} \log \lambda_{ij}(\tau).
\end{align*}

\idea{

  Following the same approach as in the single-entity case, the negative log-likelihood of the observed counting measures is given by
  \begin{align*}
    \Lcal(\{\Ybb_{ij}\}; U, V, \{f_r\})
     & = \sum_{i=1}^N \sum_{j=1}^N \Bigg(
    \int_{\Tcal} \sum_{r=1}^R u_{ir} V_{jr} f_r(t)\,dt                                 \\
     & -\int_{\Tcal} \log\left(\sum_{r=1}^R u_{ir} V_{jr} f_r(t)\right)\,d\Ybb_{ij}(t)
    \Bigg).
  \end{align*}

  And the gradients can be derived similarly as in the previous section.

  We should write the loss terms l_{ij} and then write the gradient multiplicative updates :
  - Grad l_ij wrt u_i
  - Grad l_ij wrt v_j
  - Grad l_ij wrt Gamma}

\subsection{Gradients for the Dynamic Network Case\label{appendix:dynamic_networks}}
In the following, we derive the gradients of the loss function with respect to the parameters $U$, $V$ and $\Gamma$ in the case of dynamic networks.
We denote $\otimes$ the outer product, and $\odot$ the element-wise product. We recall that the intensity function for pair $(i,j)$ is given by $\lambda_{ij}(t) = u_i \odot v_j \cdot \Gamma \Phi(t)$, where $u_i$ and $v_j$ are the factor loadings for entities $i$ and $j$, $\Gamma$ is the matrix of basis coefficients, and $\Phi(t)$ is the vector of basis functions.
  \begin{align*}
    \nabla_{u_i} \lambda_{ij}(t)
    &= v_j \odot \Gamma \Phi(t)\\
    \nabla_{v_j} \lambda_{ij}(t)
    &= u_i \odot \Gamma \Phi(t)\\
    \nabla_{\Gamma} \lambda_{ij}(t)
    &= u_i \odot v_j \otimes \Phi(t)
  \end{align*}

  The gradient of the cumulant term $\int \lambda_{ij}(t) dt$ is then straightforward by linearity of the integral. For the second term, we use the chain rule:
  \begin{align*}
    \nabla_{x} \log(\lambda_{ij}(\tau)) = \frac{\nabla_{x} \lambda_{ij}(\tau)}{\lambda_{ij}(\tau)}
  \end{align*}
  where $x$ is either $u_i$, $v_j$ or $\Gamma$.

Denoting $\Phi(t) = (\phi_1(t), \ldots, \phi_B(t))^T$ and $I(\Phi) = \int_{\Tcal} \Phi(t) dt$, the full gradients of the loss $l_{ij}$ write respectively

\begin{align*}
  \nabla_{u_i} l_{ij}
   & =
  v_j\odot \Gamma \cdot I(\Phi) - \sum_{\tau \in \Ecal_{ij}} \frac{v_j \odot \Gamma \Phi(\tau)}{u_i \odot v_j \cdot \Gamma \Phi(\tau)}          \\
  \nabla_{v_j} l_{ij}
   & =
  u_i\odot \Gamma \cdot I(\Phi) - \sum_{\tau \in \Ecal_{ij}} \frac{u_i \odot \Gamma \Phi(\tau)}{u_i \odot v_j \cdot \Gamma \Phi(\tau)}          \\
  \nabla_{\Gamma} l_{ij}
   & =u_i \odot v_j \otimes I(\Phi) - \sum_{\tau \in \Ecal_{ij}} \frac{u_i \odot v_j \otimes \Phi(\tau)}{u_i \odot v_j \cdot \Gamma \Phi(\tau)}
\end{align*}

Optimization can be done using multiplicative updates or ALS as in the single-entity case, with the corresponding updates for $u_i$, $v_j$, and $\Gamma$.
The full algorithm is given in algorithm \ref{alg:dynamic-eventnmf}.

\begin{algorithm}
  \caption{Multiplicative updates for dynamic network EventNMF}
  \label{alg:dynamic-eventnmf}
  \begin{algorithmic}[1]
    \REQUIRE Event sets $\{\Ecal_{ij}\}_{i,j=1}^N$, basis functions $\{\phi_b\}_{b=1}^B$, rank $R$, number of iterations $K$
    \ENSURE Nonnegative factors $U\in\Rbb_+^{N\times R}$, $V\in\Rbb_+^{N\times R}$ and $\Gamma\in\Rbb_+^{R\times B}$
    \STATE Initialize $U$, $V$ and $\Gamma$ with positive entries
    \FOR{$k=1,\dots,K$}
    \FOR{$i,j=1,\dots,N$}
    \STATE Compute $\lambda_{ij}(\tau)=u_i \odot v_j \cdot \Gamma \Phi(\tau)$ for all $\tau\in\Ecal_{ij}$
    \FOR{$r=1,\dots,R$}
    \STATE Update
    \[
      u_{ir}\leftarrow u_{ir}
      \frac{\sum_{\tau\in\Ecal_{ij}}\frac{v_{jr}\sum_{b=1}^B \gamma_{rb}\phi_b(\tau)}{\lambda_{ij}(\tau)}}
      {v_{jr}\sum_{b=1}^B \gamma_{rb} I(\phi_b)}
    \]
    \[
      v_{jr}\leftarrow v_{jr}
      \frac{\sum_{\tau\in\Ecal_{ij}}\frac{u_{ir}\sum_{b=1}^B \gamma_{rb}\phi_b(\tau)}{\lambda_{ij}(\tau)}}
      {u_{ir}\sum_{b=1}^B \gamma_{rb} I(\phi_b)}
    \]
    \ENDFOR
    \ENDFOR
    \FOR{$r=1,\dots,R$}
    \FOR{$b=1,\dots,B$}
    \STATE Update
    \[
      \gamma_{rb}\leftarrow \gamma_{rb}
      \frac{\sum_{i,j=1}^N u_{ir} v_{jr}\sum_{\tau\in\Ecal_{ij}}\frac{\phi_b(\tau)}{\lambda_{ij}(\tau)}}
      {\sum_{i,j=1}^N u_{ir} v_{jr} I(\phi_b)}
    \]
    \ENDFOR
    \ENDFOR
    \ENDFOR
  \end{algorithmic}
\end{algorithm}

\section{Synthetic Experiment Details}

\subsection{Synthetic Data Generation\label{appendix:synthetic_data}}

\begin{figure}[h!]
  \centering
  \begin{subfigure}[t]{0.3\columnwidth}
    \centering
    \includegraphics[width=\textwidth]{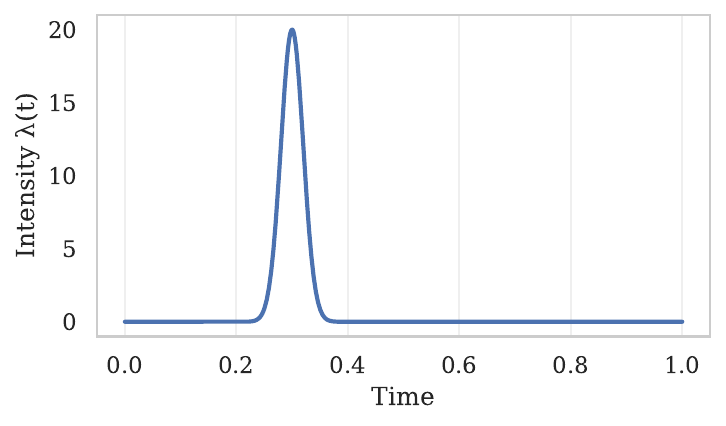}
    \caption{$f_1(t)$}
    \label{fig:synthetic-group-1}
  \end{subfigure}
  \begin{subfigure}[t]{0.3\columnwidth}
    \centering
    \includegraphics[width=\textwidth]{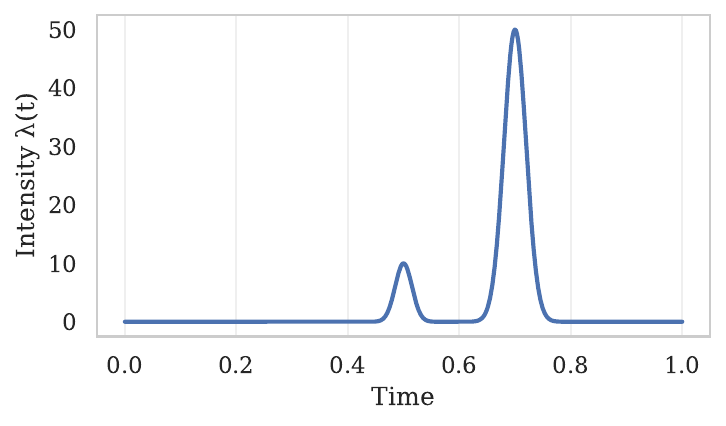}
    \caption{$f_2(t)$}
    \label{fig:synthetic-group-2}
  \end{subfigure}
  \begin{subfigure}[t]{0.3\columnwidth}
    \centering
    \includegraphics[width=\textwidth]{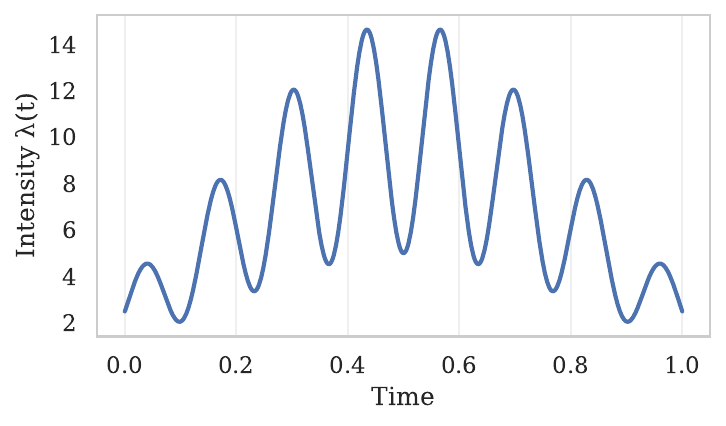}
    \caption{$f_3(t)$}
    \label{fig:synthetic-group-3}
  \end{subfigure}
  \caption{Ground truth temporal factors for the synthetic data experiment. }
  \label{fig:synthetic-ground-truth}
\end{figure}

We generate $N$ entities divided into $R=3$ groups over the time horizon $T=1$. Each entity's events are drawn from an inhomogeneous Poisson process with the intensity function of its group: a Gaussian bump ($f_1$), a double-peaked function ($f_2$), and a sinusoidally modulated envelope ($f_3$), illustrated in Figure~\ref{fig:synthetic-ground-truth}. The true entity loadings are one-hot group indicators, so the data has exact rank-$3$ structure.

From left to right
\begin{itemize}
  \item $f_1(t)= 20 \exp\left(-\frac{(t - 0.3)^2}{2 \cdot 0.02^2}\right)$is a smooth Gaussian bump, 
  \item $f_2(t) = 10 \exp\left(-\frac{(t - 0.5)^2}{2 \cdot 0.015^2}\right) + 50 \exp\left(-\frac{(t - 0.7)^2}{2 \cdot 0.02^2}\right)$ is a double-peaked function, and
  \item $f_3(t) = 5 \exp\left(-\frac{(t - 0.5)^2}{2 \cdot 0.3^2}\right) \left(1 + 0.5 \sin(15 \pi t)\right)$ is a sinusoidally modulated envelope.
\end{itemize}

\subsection{Evaluation Metrics}
\idea{\subsection{Evaluation Metrics}
\label{app:metrics}

Given estimated intensities $\hat{\lambda}_i(t)$ and observed events $\Ecal_i$, the three metrics are defined as follows. The \emph{negative log-likelihood} (NLL) is
\[
\text{NLL} = -\frac{1}{N}\sum_i \left( \sum_{\tau\in\Ecal_i} \log \hat{\lambda}_i(\tau) - \int_0^T \hat{\lambda}_i(t)\,dt \right),
\]
computed separately on train and test events, where test events are obtained by Bernoulli thinning with $p_{\mathrm{train}}=0.8$. The \emph{normalized mean squared error} (NMSE) between true and estimated intensities is
\[
\text{NMSE} = \frac{\sum_i \int_0^T (\lambda_i(t) - \hat{\lambda}_i(t))^2\,dt}{\sum_i \int_0^T \lambda_i(t)^2\,dt}.
\]
The \emph{normalized factor error} (NFISE) between true factors $f_r(t)$ and recovered factors $\hat{f}_r(t)$, after optimal permutation alignment, is
\[
\text{NFISE} = \frac{\min_{\sigma} \frac{1}{R}\sum_r \int_{0}^{T}(f_r(t) - \hat{f}_{\sigma(r)}(t))^2\,dt}{\frac{1}{R}\sum_r \int_{0}^{T} f_r(t)^2\,dt},
\]
where $\sigma$ ranges over all permutations of $\{1,\ldots,R\}$.
}
\label{app:metrics}

Given estimated intensities $\hat{\lambda}_i(t)$ and observed events $\Ecal_i$, the three metrics are defined as follows. The \emph{negative log-likelihood} (NLL) is
\[
\text{NLL} = -\frac{1}{N}\sum_i \left( \sum_{\tau\in\Ecal_i} \log \hat{\lambda}_i(\tau) - \int_0^T \hat{\lambda}_i(t)\,dt \right),
\]
computed separately on train and test events, where test events are obtained by Bernoulli thinning with $p_{\mathrm{train}}=0.8$. The \emph{normalized mean squared error} (NMSE) between true and estimated intensities is
\[
\text{NMSE} = \frac{\sum_i \int_0^T (\lambda_i(t) - \hat{\lambda}_i(t))^2\,dt}{\sum_i \int_0^T \lambda_i(t)^2\,dt}.
\]
The \emph{normalized factor error} (NFISE) between true factors $f_r(t)$ and recovered factors $\hat{f}_r(t)$, after optimal permutation alignment, is
\[
\text{NFISE} = \frac{\min_{\sigma} \frac{1}{R}\sum_r \int_{0}^{T}(f_r(t) - \hat{f}_{\sigma(r)}(t))^2\,dt}{\frac{1}{R}\sum_r \int_{0}^{T} f_r(t)^2\,dt},
\]
where $\sigma$ ranges over all permutations of $\{1,\ldots,R\}$.

\subsection{Computational Setup and Scalability}
\label{app:scalability}

All experiments were run on a server equipped with an Intel Xeon Gold 6136 CPU (3.00\,GHz, 48 cores) and 1\,TB of RAM. No GPU was used; all computations are CPU-based.

\begin{figure}[h!]
  \centering
  \includegraphics[width=0.55\columnwidth]{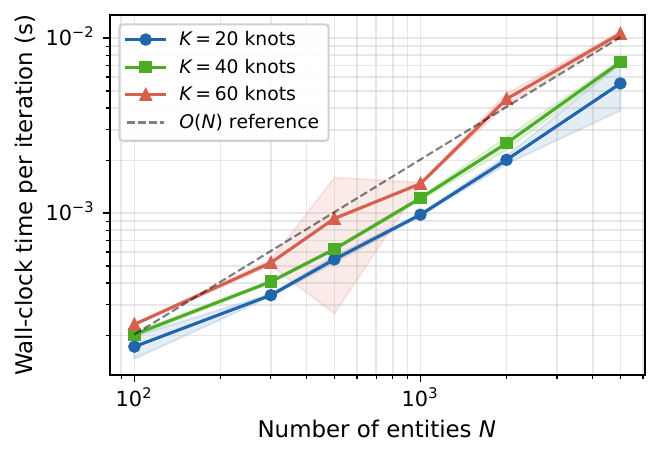}
  \caption{Wall-clock time per multiplicative-update iteration as a function of the number of entities $N$, for three choices of basis size $B$. The dashed line shows an $O(N)$ reference. Timing was measured on an Intel Xeon Gold 6136 CPU (3.00\,GHz).}
  \label{fig:timing}
\end{figure}

The per-iteration cost of the multiplicative update algorithm scales as $O(N \cdot \bar{E} \cdot B)$, where $N$ is the number of entities, $\bar{E}$ is the average number of events per entity, and $B$ is the number of basis functions. Figure~\ref{fig:timing} shows median wall-clock time \emph{per iteration} across a range of $N$ and $B$ values (rank $R=3$, time horizon $T=1$, medians over 3 repeats). Runtimes scale approximately linearly in $N$, consistent with the theoretical complexity and the $O(N)$ reference line. At $N=1000$ with $B=40$ basis functions, one iteration takes approximately 1.2\,ms; at $N=5000$ the same takes approximately 7\,ms.

